\documentclass[journal]{IEEEtran}

\usepackage{graphicx} 
\usepackage{amsfonts}
\usepackage{amssymb}
\usepackage{amsmath}
\usepackage{amsthm}
\usepackage{xfrac}
\usepackage{algorithm}
\usepackage{algorithmic}
\usepackage{mathtools}
\usepackage{subfig}
\usepackage{rotating}
\usepackage[table]{xcolor}
\definecolor{light-gray}{gray}{0.9}

\theoremstyle{definition}

\newtheorem{lemma}{Lemma}[]
\newtheorem{remark}{Remark}[]

\usepackage{cite}

\ifCLASSINFOpdf
\else
\fi

\begin{document}
%
\title{Supervised Hyperalignment\\ for multi-subject fMRI data alignment}
%
%
%

\author{Muhammad Yousefnezhad, Alessandro Selvitella, Liangxiu Han, Daoqiang Zhang
\thanks{M. Yousefnezhad and D. Zhang are with the College of Computer Science and Technology, Nanjing University of Aeronautics and Astronautics, Nanjing 210016, China (e-mail: dqzhang@nuaa.edu.cn).}
\thanks{M. Yousefnezhad and A. Selvitella are with the Deparment of Computing Science, University of Alberta, Edmonton T6G 2R3, AB, Canada (email: myousefnezhad@ualberta.ca)}
\thanks{A. Selvitella is with the Department of Mathematical Sciences, Purdue University Fort Wayne, 2101 E Coliseum Blvd, Fort Wayne, 46805 United States (email: aselvite@pfw.edu)}
\thanks{L. Han is with the School of Computing, Mathematics and Digital Technology, Manchester Metropolitan University, Manchester M156BH U.K. (e-mail: l.han@mmu.ac.uk) }
\thanks{D. Zhang serves as the corresponding author.}
}


%
%

\markboth{IEEE Transactions on Cognitive and Developmental Systems}{}%
%



\maketitle

\begin{abstract}
Hyperalignment has been widely employed in Multivariate Pattern (MVP) analysis to discover the cognitive states in the human brains based on multi-subject functional Magnetic Resonance Imaging (fMRI) datasets. Most of the existing HA methods utilized unsupervised approaches, where they only maximized the correlation between the voxels with the same position in the time series. However, these unsupervised solutions may not be optimum for handling the functional alignment in the supervised MVP problems. This paper proposes a Supervised Hyperalignment (SHA) method to ensure better functional alignment for MVP analysis, where the proposed method provides a supervised shared space that can maximize the correlation among the stimuli belonging to the same category and minimize the correlation between distinct categories of stimuli. Further, SHA employs a generalized optimization solution, which generates the shared space and calculates the mapped features in a single iteration, hence with optimum time and space complexities for large datasets. Experiments on multi-subject datasets demonstrate that SHA method achieves up to 19\% better performance for multi-class problems over the state-of-the-art HA algorithms.
\end{abstract}



%
\IEEEpeerreviewmaketitle

\sloppy

\section{Introduction}
\IEEEPARstart{B}{rain} decoding, which is a conjunction between neuroscience and machine learning, extracts meaningful patterns (signatures) from neural activities of the human brain. Most of the brain decoding approaches employed functional Magnetic Resonance Imaging (fMRI) technology for visualizing the brain activities because it can provide better spatial resolution in comparison with other imaging techniques \cite{haxby11,haxby14,diedrichsen17,xu12,lorbert12,chen14,chen15,tony16,tony17,tony17a}. fMRI can be used as a proxy to illustrate the brain neural activities by analyzing the Blood Oxygen Level Dependent (BOLD) signals \cite{tony16,tony17,diedrichsen17}. As one of the most popular supervised techniques in fMRI analysis, Multivariate Pattern (MVP) classification can map neural activities to distinctive brain tasks \cite{haxby11,chen15,oswal16}. MVP can generate a classification (cognitive) model, i.e., decision surfaces \cite{haxby14,wang2018a,wang2018b}, in order to predict patterns associated with different cognitive states \cite{mohr15,tony17a,haxby14,diedrichsen17}. This model can help us to figure out how the human brain works \cite{haxby11,haxby14}. MVP analysis has an extensive range of applications to seek novel treatments for mental diseases \cite{haxby14,tony17a}.

As a fundamental challenge in supervised fMRI studies, the generated MVP models must be generalized and validated across subjects \cite{haxby11,haxby14,xu12,lorbert12,chen14,chen15,tony17a,diedrichsen17}. However, neuronal activities in multi-subject fMRI dataset must be aligned to improve the performance of the final results \cite{haxby11,chen14}. Technically, there are two different kinds of alignment techniques that can be used in harmony, i.e., anatomical alignment and functional alignment \cite{haxby11,chen14,tony17a}. The anatomical alignment as a general preprocessing step in fMRI analysis aligns the brain patterns by using anatomical features, which is extracted from structural MRI in the standard space (Talairach \cite{talairach88} or Montreal Neurological Institute (MNI) \cite{mazziotta01}). Nevertheless, the performance of anatomical alignment techniques are limited based on the shape, size, and spatial location of functional loci differ across subjects \cite{watson93,rademacher93, tony17a}. In contrast, the functional alignment can directly align the neural activities across subjects, which has been widely used in fMRI studies.

Most of recent studies in functional alignment \cite{haxby11,haxby14,xu12,lorbert12,chen14,chen15,chen16,tony17a} have used Hyperalignment (HA) \cite{haxby11}. HA refers to the feature alignment of multi-subject data, where a shared space is generated from neural activities across subjects, and then the mapped features can be utilized by MVP techniques in order to boost the performance of the classification analysis. In practice, HA applies a Generalized Canonical Correlation Analysis (GCCA) approach (aka multi-set CCA) to temporally-aligned neural activities across subjects, where a unique time point must represent the same simulation for all subjects \cite{xu12,lorbert12,chen14,diedrichsen17}. We recently illustrated that the performance of HA methods might not be optimum for supervised fMRI analysis (i.e., MVP problems) because they mostly employed unsupervised GCCA techniques for aligning the neural activities across subjects \cite{tony17a}. Therefore, we have developed Local Discriminant Hyperalignment (LDHA) \cite{tony17a} for improving the alignment accuracy in the MVP problems. Although LDHA can improve the performances of both functional alignment and MVP analysis, its objective function cannot directly calculate a supervised shared space and still uses the classical unsupervised shared space \cite{tony17a}. Thus, it cannot provide stable performance and acceptable runtime for large datasets in the real-world applications.

As the main contribution, this paper introduces a new supervised functional alignment method for MVP analysis that is called Supervised Hyperalignment (SHA), which provides a generalized optimization solution.  In a nutshell, SHA generates a supervised shared space where each stimulus is only compared with the share space to align neural activities and calculates mapped feature in a single iteration. As a result, SHA can maximize the correlation among the stimuli belonging to the same category and minimize the correlation between distinct categories of stimuli. SHA utilizes a generalized solution for optimization, and the training-set is not referenced in the testing stage. The proposed method has an optimum time and space complexities for large datasets.

The rest of this paper is organized as follows: Section II represents related work. In Section III, a brief overview of HA methods is introduced; Section IV details the proposed Supervised Hyperalignment; Section V reported the experimental results; Section VI concludes this research and highlights the future work.


\section{Related Works}
Several none-HA based studies used functional and anatomical features for aligning fMRI datasets across subjects. Conroy et al. proposed a new method to maximize the Inter-Subject Correlation (ISC) by utilizing the cortical warping \cite{conroy09}. Sabuncu et al. employed both within and across the cortical area for maximizing the ISC across subjects \cite{sabuncu10}. Micheal et al. developed a functional alignment for rest-mode fMRI (rs-fMRI) by utilizing Group Independent Component Analysis (GICA) and Independent Vector Analysis (IVA) algorithms. Since time synchronized stimulus is not assumed in this method, data is concatenated across the time dimension, and then the independent components are learned \cite{michael15}. Langs et al. used functional alignment techniques to distinguish healthy subjects and the subjects with tumors in fMRI datasets \cite{Langs10}. 

Haxby et al. developed Hyperalignment (HA) as an `anatomy free' method based on functional features \cite{haxby11}. Technically, HA utilized the Procrustean transformation \cite{schonemann66} to map the neural activities of different subjects into a high-dimensional shared model (template). HA can rapidly increase the performance of MVP analysis in comparison with the methods that just utilized the anatomical alignment \cite{haxby11,haxby14}. Xu et al. developed the Regularized Hyperalignment (RHA) \cite{xu12} by reformulating HA as a Canonical Correlation Analysis (CCA) \cite{gower04} problem. Dmochowski et al. applied correlated component analysis to maximize ISC by aggregating the subjects’ data into an individual matrix \cite{dmochowski12}. As a nonlinear kernel approach, Lorbert et al. proposed Kernel Hyperalignment (KHA) for mapping the neural activities into an embedding space \cite{lorbert12}. Further, Yousefnezhad et al. developed Deep Hyperalignment (DHA) \cite{Tony17NIPS} that replaces a deep parametric kernel rather than the standard non-parametric kernel approaches in KHA. In practice, this method can find a custom non-linear space for each subject and then align the neural activities form this non-linear space to a shared space \cite{Tony17NIPS}. Sui et al. \cite{sui11,sui13} combined multimodal CCA and joint Independent Component Analysis (ICA) techniques in order to identify the unique and shared spaces in the multimodal data. Chen et al. \cite{chen14} introduced a two-phase joint Singular Value Decomposition Hyperalignment (SVDHA) algorithm, where Singular Value Decomposition (SVD) is used for reducing data dimensions. And then, HA aligns the neural responses across subjects in the lower dimensional feature space. 

As an ensemble approach, Guntupalli et al. proposed SearchLight Hyperalignment (SLHA) that is quasi-CCA models fit on patches of the neural activities. Indeed, SLHA applies functional alignment within searchlights, and then the aligned data in local transformations are aggregated into a global transformation. In other words, SLHA can constrain the anatomical location of the neural activities across subjects \cite{guntupalli16}. Chen et al. introduced Shared Response Model (SRM) for combining multi-subject fMRI images, which accounts for distinctive functional topographies among anatomically aligned fMRI images. SRM is technically equivalent to Probabilistic CCA \cite{chen15}. Turek et al. proposed a Semi-Supervised SRM (SS-SRM) that simultaneously applies the alignment and performs the analysis. This method still used the unsupervised SRM \cite{chen15} for alignment, which were based on unsupervised features ($\mathcal{L}_{SRM}$ in \cite{turek16}) and the semi-supervised Multinomial Logistic Regression for Classification Analysis \cite{turek16}. Yousefnezhad et al. developed LDHA, where this method applied the locality information (i.e., the nearest neighbors of both within-class and between-classes neural activities) for improving the performance of classification \cite{tony17a}.

\begin{figure*}[!t]
	\begin{center}
		\includegraphics[width=0.98\textwidth,height=0.38\linewidth]{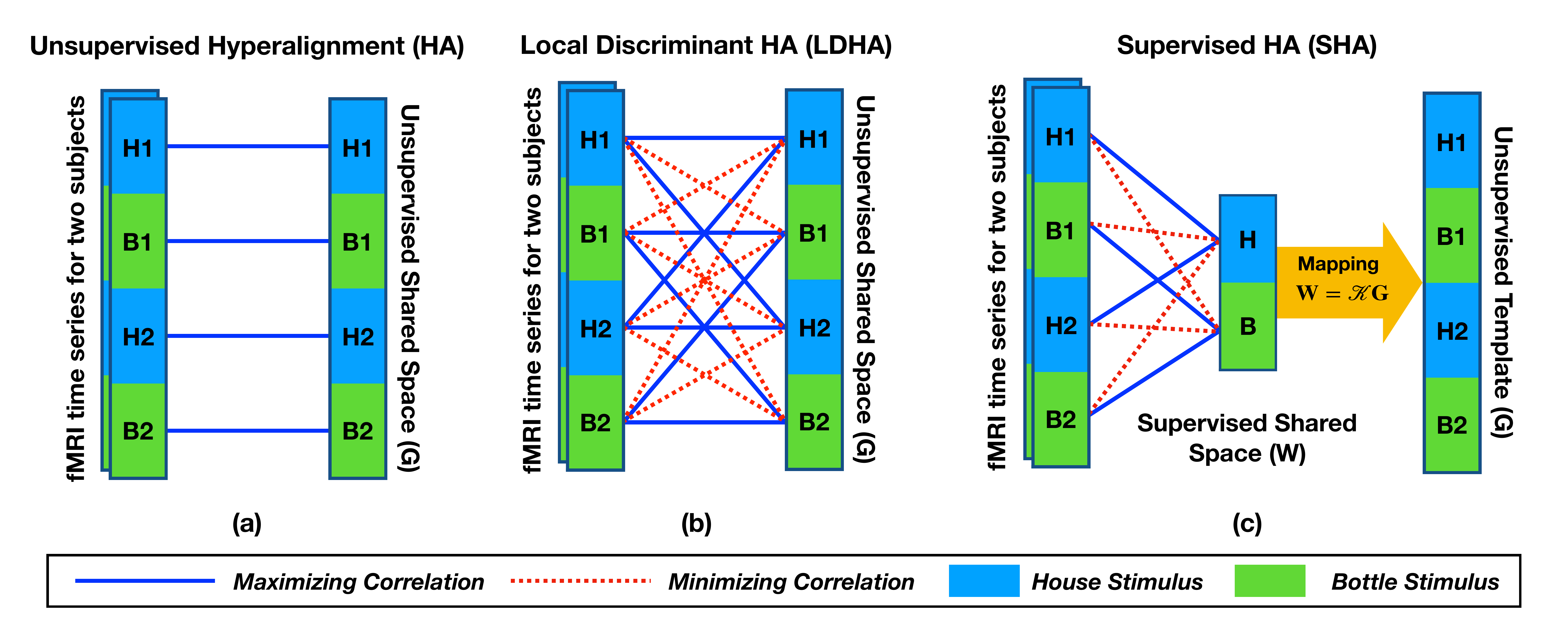}\\
		\caption{Comparison of different HA algorithms for aligning neural activities.}
		\label{fig:DifferentHAs}
	\end{center}
	\vskip -0.2in
\end{figure*} 

\section{Hyperalignment Background}
A fMRI dataset can be defined by $\mathbf{X}^{(i)} \in \mathbb{R}^{T \times V}\text{, } i=1\text{:}S$, where $S$ is the number of subjects, $V$ denotes the number of voxels, and $T$ is the number of time points in units of Time of Repetitions (TRs). For simplicity, this paper considers $\mathbf{X}^{(i)} \sim \mathcal{N}(0,1)$ which is normalized by zero-mean and unit-variance in the preprocessing step. It is worth noting that $(\mathbf{X}^{(i)})^\top \mathbf{X}^{(j)}$ as the correlation map is not generally full rank because the number of voxels is more than TRs in most of datasets \cite{tony17a}. Like previous studies \cite{chen14,conroy09,lorbert12,xu12,tony17a}, $\mathbf{X}^{(i)} \text{, } i=1\text{:}S$ is time synchronized to provide temporal alignment. Here, each time point demonstrates the same stimuli for all subjects \cite{xu12,lorbert12}. In fact, the columns of $\mathbf{X}^{(i)}$ are aligned across subjects by utilizing HA methods \cite{conroy09,xu12}. The original HA can be defined as follows where $tr()$ is the trace function:  \cite{haxby11,xu12} 
\begin{equation}\label{eq:HAMax}
\begin{split}
\underset{\mathbf{R}^{(i)}, \mathbf{R}^{(j)}}{\max}\Bigg\{\sum_{i = 1}^{S}\sum_{j = i+1}^{S}\text{tr}\Big(\big(\mathbf{R}^{(i)}\big)^\top\mathbf{C}^{(i,j)}\mathbf{R}^{(j)}\Big)\Bigg\},\\
\text{s.t. \quad}\big(\mathbf{R}^{(\ell)}\big)^\top\mathbf{C}^{(\ell,\ell)}\mathbf{R}^{(\ell)}=\mathbf{I}_{V}\text{, } \ell=1\text{:}S,
\end{split}
\end{equation}
where $\mathbf{I}_{V}$ is a $V\times V$ identity matrix, $\mathbf{R}^{(\ell)}\in\mathbb{R}^{V \times V}\text{, }\ell=1\text{:}S$ denotes the hyperalignment mappings, and the covariance matrices $\mathbf{C}^{(i,j)}=(\mathbf{X}^{(i)})^\top \mathbf{X}^{(j)}$, $\mathbf{C}^{(i,j)} \in \mathbb{R}^{V \times V}$ are positive definite and symmetric. Since functional alignment techniques consider that the mappings $\mathbf{R}^{(\ell)}, \ell=1\text{:}S$ must find the noisy `rotations' of a shared space \cite{haxby11}, the constraint $\big(\mathbf{R}^{(\ell)}\big)^\top\mathbf{C}^{(\ell,\ell)}\mathbf{R}^{(\ell)}=\mathbf{I}_{V}\text{, } \ell=1\text{:}S$ can avoid overfitting issues \cite{xu12,lorbert12}.

\begin{remark}\label{rm:MatrixA}\emph{
		If $\mathbf{C}^{(\ell,\ell)} = \mathbf{I}_{V}$, then equation \eqref{eq:HAMax} can be reformulated as a multi-set orthogonal Procrustes problem that generally is employed in shared analysis. By considering $\mathbf{C}^{(\ell,\ell)} = (\mathbf{X}^{(\ell)})^\top \mathbf{X}^{(\ell)}$, equation \eqref{eq:HAMax} is equivalent to multi-set Canonical Correlation Analysis (CCA) \cite{lorbert12,xu12,chen14,tony17a}. }
\end{remark}

As mentioned before, equation \eqref{eq:HAMax} may not be optimum for the classification analysis. In order to improve the performance of functional alignment, Local Discriminant Hyperalignment (LDHA) uses following objective function \cite{tony17a}:
\begin{equation}\label{eq:LDHA}
\begin{split}
\underset{\mathbf{R}^{(i)}, \mathbf{R}^{(j)}}{\max}\Bigg\{\sum_{i = 1}^{S}\sum_{j = i+1}^{S}\text{tr}\Big(\big(\mathbf{R}^{(i)}\big)^\top\big(\mathbf{\Delta}^{(i,j)} - \eta\mathbf{\Omega}^{(i,j)}\big)\mathbf{R}^{(j)}\Big)\Bigg\},
\end{split}
\end{equation}
where $\eta$ is the scaling factor that can be calculated by using the number of within-class elements over the square of all time points. Further, the covariance within-class matrix $\mathbf{\Delta}^{(i,j)}=\Big\{\mathbf{\delta}^{(i,j)}_{mn}\Big\}$ and the covariance between-class matrix $\mathbf{\Omega}^{(i,j)}=\Big\{\mathbf{\omega}^{(i,j)}_{mn}\Big\}$ are denoted as follows:
\begin{equation}\label{eq:withinclass}
\mathbf{\delta}^{(i,j)}_{mn} = \sum_{\ell=1}^{T}\sum_{k=1}^{T} \mathbf{\alpha}_{\ell k}\mathbf{x}_{\ell m}^{(i)}\mathbf{x}_{kn}^{(j)} + \mathbf{\alpha}_{\ell k}\mathbf{x}_{\ell n}^{(i)}\mathbf{x}_{k m}^{(j)},
\end{equation}
\begin{equation}\label{eq:betweenclasses}
\mathbf{\omega}^{(i,j)}_{mn} = \sum_{\ell=1}^{T}\sum_{k=1}^{T} (1 - \mathbf{\alpha}_{\ell k})\mathbf{x}_{\ell m}^{(i)}\mathbf{x}_{kn}^{(j)} + (1 - \mathbf{\alpha}_{\ell k})\mathbf{x}_{\ell n}^{(i)}\mathbf{x}_{k m}^{(j)},
\end{equation}
where $\mathbf{\alpha}_{\ell k}=1$ for within-class elements, otherwise it is zero. Moreover, $\mathbf{x}^{(i)}_{mn} \in \mathbb{R}$ is the neural activity for the $m\text{-}th$ time point ($m=1\text{:}T$) and the $n\text{-}th$ voxel ($n=1\text{:}V$) in the $i\text{-}th$ subject \cite{tony17a}. 
\begin{figure*}[!t]
	\begin{center}
		\includegraphics[width=0.67\textwidth]{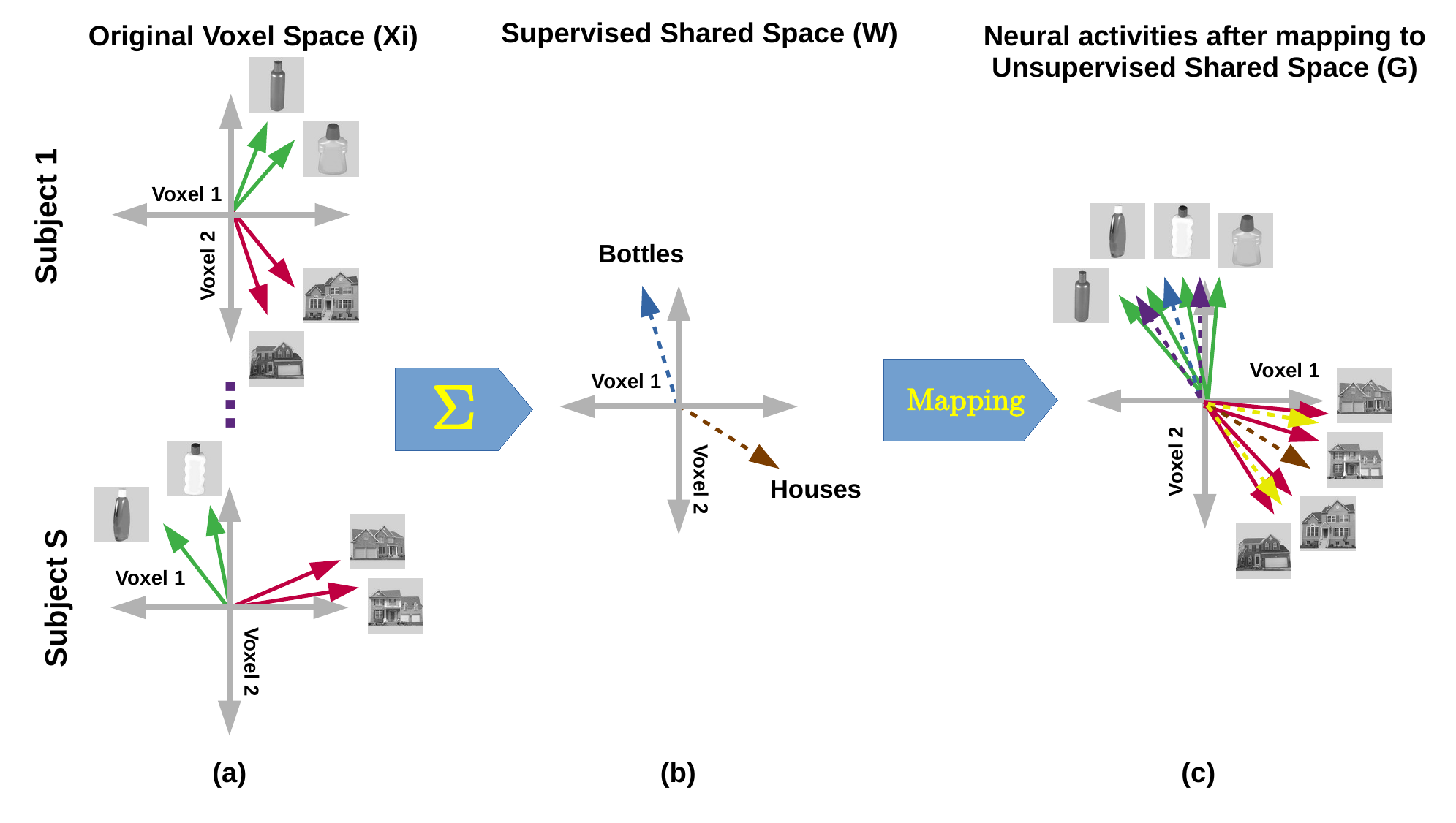}\\ 
		\caption{An example of a) original space of neural activities; b) supervised shared space; c) unsupervised shared space}
		\label{fig:SHA}
	\end{center}
	\vskip -0.3in
\end{figure*}

\section{The Proposed Supervised Hyperalignment}
Figure \ref{fig:DifferentHAs} compares the \textbf{\emph {main difference}} between the proposed SHA and other HA approaches. As depicted in this figure, two subjects watch two photos of houses as well as two photos of bottles, where $[\mathbf{H1}, \mathbf{B1}, \mathbf{H2}, \mathbf{B2}]$ shows the sequence of stimuli. Here, the shared spaces can be calculated by employing different correlations between neural activities. Figure \ref{fig:DifferentHAs}.a demonstrates that the unsupervised HA just maximizes the correlation between the voxels with the same position in the time series because it does not use the supervision information.

Figure \ref{fig:DifferentHAs}.b illustrates the LDHA approach, where it utilizes the unsupervised shared space for the alignment problem. Indeed, the main issue in equation \eqref{eq:LDHA} is that the covariance matrices ($\mathbf{\Delta}^{(i,j)} - \eta\mathbf{\Omega}^{(i,j)}$) cannot decompose to the product of a symmetric matrix. In order to calculate the shared space in LDHA, each pair of stimuli must be separately compared with each other, and the shared space is gradually updated in each comparison (see Algorithm 2 in \cite{tony17a}). Therefore, LDHA cannot use a generalized optimization approach (such as GCCA) and its time complexity is not efficient for large datasets.

As shown in Figure \ref{fig:DifferentHAs}.c, supervised hyperalignment (SHA) consists of two main steps:
\begin{enumerate}
	\item  Generating a supervised shared space, where each stimulus is only compared with the shared space to align the neural activities;
	\item Calculating the mapped features in a single iteration. 
\end{enumerate}

Figure \ref{fig:SHA} also explains the geometry of the presented example in Figure \ref{fig:DifferentHAs}.c, where SHA maps the multi-subject neural activities to a supervised shared space and then mapped features are transformed into an unsupervised shared space for using in the testing phase. As Figure  \ref{fig:SHA}.b depicted, a \textbf{\emph{supervised shared space}} $\mathbf{W} \in \mathbb{R}^{L \times V}$ refers to a vector space, where each category of stimuli has a unique neural signature across all subjects. These signatures are generated by considering all within-class and between-class relations. As Figure \ref{fig:SHA}.c illustrated, an \textbf{\emph{unsupervised shared space}} $\mathbf{G} \in \mathbb{R}^{T \times V}$ is a vector space, where $i\text{-}th$ row is the shared neural activities belong to the $i\text{-}th$ stimulus (time point). In practice, \emph{each neural signature in the supervised space represents an abstract concept such as houses or bottles, while a mapped neural activity in the unsupervised shared space belongs to a unique stimulus such as the big white house or the gray bottle.} As an example, we have two groups of stimuli for each subject (depicted by red and green arrows) in Figure \ref{fig:SHA}. Firstly, SHA generates a supervised shared neural signature across all subject for each category of stimuli that are shown by blue (for bottles) and brown (for houses) in this figure. Finally, SHA calculates the unsupervised shared neural activities (i.e., the purple arrows for bottles and yellow arrows for houses) and then the original neural activities are mapped to this space. Our proposed algorithm is detailed in the following sections.

\subsection{Generating A Supervised Shared Space}
The neural activities belong to $\ell\text{-}th$ subject can be denoted by $\mathbf{X}^{(\ell)} \text{, } \ell=1\text{:}S$ (defined same as previous section) and the class labels that are denoted by $\mathbf{Y}^{(\ell)}=\Big\{{y}_{mn}^{(\ell)} \Big\}\text{, }\mathbf{Y}^{(\ell)}\in\{0,1\} ^{L\times T}\text{, } m=1\text{:}L\text{, }n=1\text{:}T\text{, }L>1$. Here, $L$ is the number of classes (stimulus categories). In order to infuse supervision information to the HA problem, this paper defines a supervised term as follows:
\begin{equation}\label{eq:SupCovKer}
\begin{split}
\mathbf{K}^{(\ell)} \in \mathbb{R}^{L \times T} = \mathbf{Y}^{(\ell)}\mathbf{H},
\end{split}
\end{equation}
where the normalization matrix $\mathbf{H}\in\mathbb{R}^{T\times T}$ is denoted as follows:
\begin{equation}\label{eq:H}
\begin{split}
\mathbf{H} = \mathbf{I}_{T} - \gamma{\mathbf{1}}_{T}, 
\end{split}
\end{equation}
where ${\mathbf{1}}_{T} \in \{1\}^{T\times T}$ denotes ones matrix in size $T$, and $\gamma$ makes a trade-off between within-class and between-class samples.

\begin{figure}[!t]
	\begin{center}
		\includegraphics[width=0.38\textwidth]{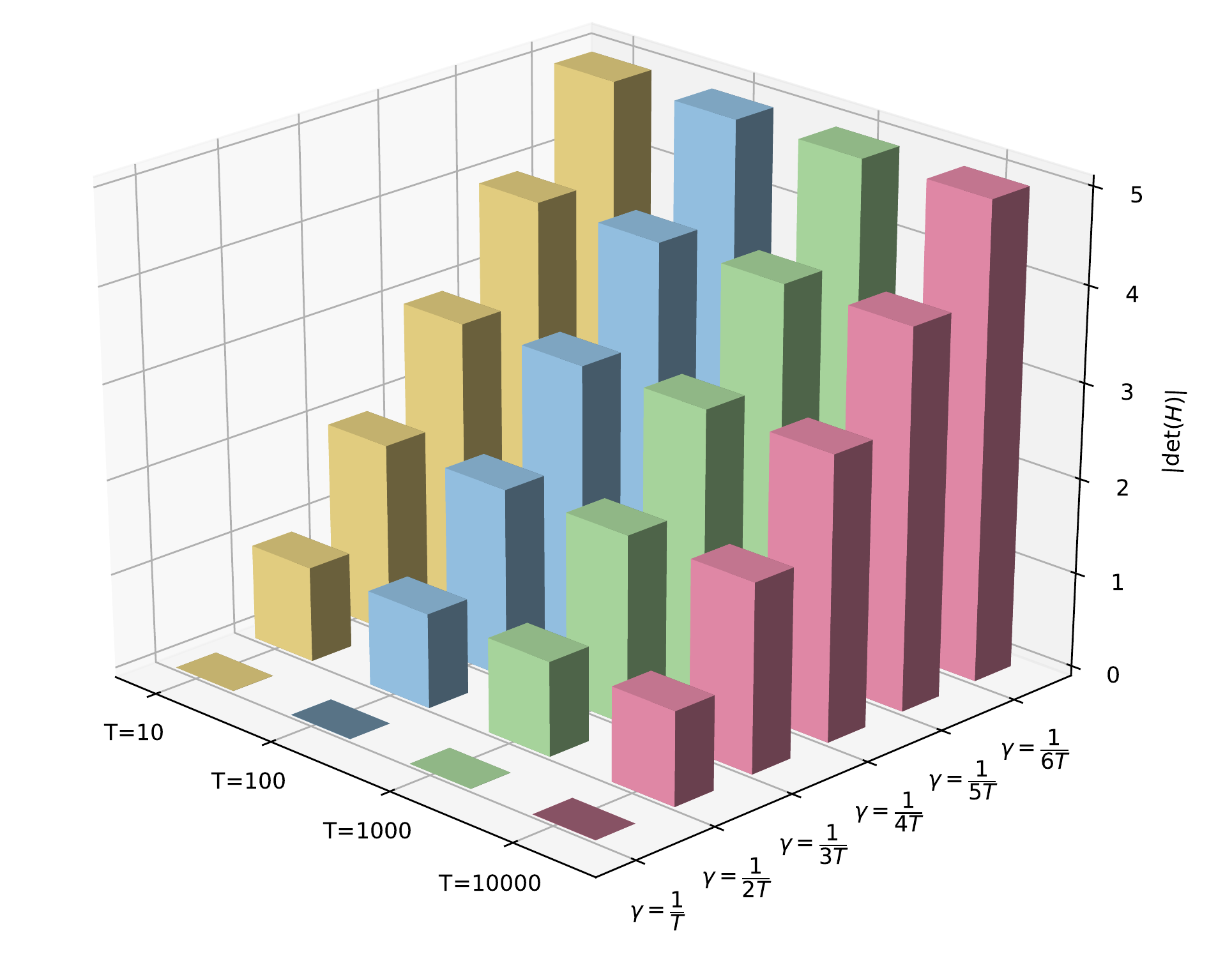}\\ 
		\caption{Comparing different supervision rates}
		\label{fig:Gamma}
	\end{center}
	\vskip -0.2in
\end{figure}

\begin{remark}\emph{
In \eqref{eq:H}, the trade-off coefficient $\gamma$ must be defined based on the number of time points. Indeed, $\gamma = 0$ removes the effect of between-class instances on each other. Further, $\gamma = \frac{1}{T}$ makes $\mathbf{H}$ idempotent, where SHA cannot find optimized solution without regularization. If we define $\gamma > \frac{1}{T}$, then the covariance matrix operates inversely, and SHA results are highly-unstable. In this paper, we are looking for $0 < \gamma < \frac{1}{T}$, where the effect of within-class instances boosts if $\gamma \to 0$. Similarity, the effect of between-class samples increases when $\gamma \to \frac{1}{T}$. Here, trade-off coefficient can be analyzed based on $\det(\mathbf{H})$. Figure \ref{fig:Gamma} shows the fluctuation of $| \det(\mathbf{H}) |$ based on different number of time points ($T$). As depicted in this figure, when $\gamma$ is assigned based on a ratio of time points, \eqref{eq:H} can produce the same trade-off for distinctive problems with different range of time points. Since this paper want to make a balance ($50/50$) trade-off between within-class and between-class samples, we employ $\gamma = \frac{1}{2T}$ for generating SHA's results.
}\end{remark}

Based on equation \eqref{eq:SupCovKer}, objective function of SHA is defined as follows: 
\begin{equation}\label{eq:SHAMax}
\begin{gathered}
\underset{\mathbf{R}^{(i)}, \mathbf{R}^{(j)}}{\max}\Bigg\{\frac{2}{S - 1}\sum_{i = 1}^{S}\sum_{j = i+1}^{S}\text{tr}\Big((\mathbf{K}^{(i)}\mathbf{X}^{(i)}\mathbf{R}^{(i)})^\top\mathbf{K}^{(j)}\mathbf{X}^{(j)}\mathbf{R}^{(j)}\Big)\\
+ \epsilon \sum_{\ell=1}^{S} \bigg\| \mathbf{R}^{(\ell)}\bigg\|_F^2\Bigg\},\\
\text{s.t. }
\big(\mathbf{R}^{(\ell)}\big)^\top\bigg(\big(\mathbf{K}^{(\ell)}\mathbf{X}^{(\ell)}\big)^\top
\mathbf{K}^{(\ell)}\mathbf{X}^{(\ell)}
+ \epsilon \mathbf{I}_V\bigg)
\mathbf{R}^{(\ell)}=\mathbf{I}_{V},
\end{gathered}
\end{equation}
where $\ell=1\text{:}S$, $\mathbf{R}^{(\ell)}\in\mathbb{R}^{V\times V}$ is the mapping matrix, and $\epsilon$ is the regularization term.

\begin{remark}\emph{By employing equation \eqref{eq:SHAMax}, the (unsupervised) regularized HA can be now considered as a special case of SHA objective function, where time points are mathematically considered independent, i.e., $\big(\mathbf{K}^{(i)}\big)^\top\mathbf{K}^{(j)} = \mathbf{I}_{T}$.
}\end{remark}

Since it is hard to find an upper bound to equation \eqref{eq:SHAMax} (evaluating the distance between current value and absolute maximum), this objective function may not be the best solution \cite{xu12,chen14}. Instead, we can rewrite equation \eqref{eq:SHAMax} as a minimization problem, where the defined objective function generates a result near to zero for an optimum solution.  
\begin{lemma}\label{lm:DHAMin}\emph{
		By switching equation \eqref{eq:SHAMax} as a minimization problem, the objective function of SHA can be written as follows, where $\ell=1\text{:}S$:}
	\begin{equation}
	\begin{gathered}\label{eq:SHA}
	\underset{\mathbf{R}^{(i)}, \mathbf{R}^{(j)}}{\min}\Bigg\{\sum_{i = 1}^{S}\sum_{j = i+1}^{S}  \bigg\| \mathbf{K}^{(i)}\mathbf{X}^{(i)}\mathbf{R}^{(i)} - \mathbf{K}^{(j)}\mathbf{X}^{(j)}\mathbf{R}^{(j)} \bigg\|^2_F\Bigg\}\text{, } \\
	\text{s.t. }
	\big(\mathbf{R}^{(\ell)}\big)^\top\bigg(\big(\mathbf{K}^{(\ell)}\mathbf{X}^{(\ell)}\big)^\top
	\mathbf{K}^{(\ell)}\mathbf{X}^{(\ell)}
	+ \epsilon \mathbf{I}_V\bigg)
	\mathbf{R}^{(\ell)}=\mathbf{I}_{V}.
	\end{gathered}
	\end{equation}
	\emph{Proof.} Please refer to Proofs section.
\end{lemma}

\begin{remark}
There are two possible approaches for solving SHA objective function. As the first approach, we can employ the optimization technique proposed in \cite{xu12} for solving \eqref{eq:SHA}. However, this solution is not unique and fast because it seeks the optimal solution by using an iterative optimization approach. In this paper, we called this method \emph{Supervised Hyperalignment with RHA optimization (SHA-R)}. Please refer to Lemma 3.2 in \cite{xu12} and then replace $\mathbf{K}^{(i)}\mathbf{X}^{(i)}$ with $\mathbf{X}_{i}$ and also $\alpha=\frac{\epsilon}{\epsilon + 1}$. As another alternative, we proposed a novel optimization solution in this paper that can find optimal parameters for  \eqref{eq:SHA} in a single iteration. In the rest of this paper, we will present the new approach and then compare the performance of SHA by using both these optimization techniques. 
\end{remark}
 
In order to solve equation \eqref{eq:SHA}, the main difficulty is that a solution for correlation objective is limited to two random variables. Therefore, a natural extension cannot be found for more than two variables \cite{rastogi15,benton17}.
 
\begin{lemma}\label{lm:GSHA}
	\emph{The equation \eqref{eq:SHA} is equivalent to:}
	\begin{equation}
	\begin{gathered}\label{eq:CCA}
	\underset{\mathbf{W}, \mathbf{R}^{(i)}}{\min}\Bigg\{\sum_{i = 1}^{S} \Big\| \mathbf{K}^{(i)}\mathbf{X}^{(i)}\mathbf{R}^{(i)} - \mathbf{W}\Big\|^2_F\Bigg\}\text{, }\\
	\text{s.t. }
	\big(\mathbf{R}^{(\ell)}\big)^\top\bigg(\big(\mathbf{K}^{(\ell)}\mathbf{X}^{(\ell)}\big)^\top
	\mathbf{K}^{(\ell)}\mathbf{X}^{(\ell)}
	+ \epsilon \mathbf{I}_V\bigg)
	\mathbf{R}^{(\ell)}=\mathbf{I}_{V}\text{, }
	\end{gathered}
	\end{equation}
	\emph{where $\ell=1\text{:}S$, and $\mathbf{W} \in \mathbb{R}^{L\times V}$ is supervised shared space:}
	\begin{equation}
	\begin{split}\label{eq:G}
	\mathbf{W} = \frac{1}{S} \sum_{j=1}^{S} \mathbf{K}^{(j)}\mathbf{X}^{(j)}\mathbf{R}^{(j)}.
	\end{split}
	\end{equation}	
	\emph{Proof.} Please see Proofs section.
\end{lemma}

\begin{lemma}\label{lm:GenMin}
\emph{This paper has used the exchangeability property of the minimization as follow: suppose function $f \in \mathbb{C}^2(\mathbb{R}^N \times \mathbb{R}^M, \mathbb{R})$ has a global minimum, then:}
{\small 
\begin{equation}\label{eq:GenMin}
\begin{gathered}	
\min_{x \in \mathbb{R}^N} \min_{y \in \mathbb{R}^M}  f(x,y)=
\min_{y \in \mathbb{R}^M} \min_{x \in \mathbb{R}^N}  f(x,y)=
\min_{(x,y) \in \mathbb{R}^N \times \mathbb{R}^M}  f(x,y).
\end{gathered}
\end{equation}}
\emph{Proof.} Please refer to Proofs section.
\end{lemma}

Based on Lemma \ref{lm:GenMin}, equation \eqref{eq:CCA} has a global optimum solution for $\mathbf{W}$ that can be independent from mapping matrices if and only if we consider following solution for each mapping matrix: $\mathbf{R}^{(\ell)}=
\Big(\big(\mathbf{K}^{(\ell)}\mathbf{X}^{(\ell)}\big)^\top\mathbf{K}^{(\ell)}\mathbf{X}^{(\ell)} + \epsilon\mathbf{I}_V\Big)^{-1}\big(\mathbf{K}^{(\ell)}\mathbf{X}^{(\ell)}\big)^\top\mathbf{W}$. In other words, the gradient equation for $\mathbf{W}$ can be solved explicitly as a function of the $\mathbf{R}^{(\ell)}$. Note that, our solution is a generalized version of the approaches proposed in \cite{rastogi15,benton17}, where we do not need to satisfy $\mathbf{W}^\top\mathbf{W}=\mathbf{I}$ because our mapping matrices are already regularized.

\begin{lemma}\label{lm:GtoA}\emph{As the objective function of SHA, equation \eqref{eq:CCA} can be rewritten as follows:}
\begin{equation}\label{eq:GtoA}
\begin{gathered}
\underset{\mathbf{W}, \mathbf{R}^{(i)}}{\min}\Bigg\{\sum_{i = 1}^{S} \Big\| \mathbf{K}^{(i)}\mathbf{X}^{(i)}\mathbf{R}^{(i)} - \mathbf{W} \Big\|^2_F\Bigg\} \equiv \underset{\mathbf{W}}{\min}\Big\{\text{tr}\big(\mathbf{W}^\top\mathbf{U}\mathbf{W}\big)\Big\},\\
\text{s.t. }\mathbf{R}^{(\ell)}=
\Big(\big(\mathbf{K}^{(\ell)}\mathbf{X}^{(\ell)}\big)^\top\mathbf{K}^{(\ell)}\mathbf{X}^{(\ell)} + \epsilon\mathbf{I}_V\Big)^{-1}\big(\mathbf{K}^{(\ell)}\mathbf{X}^{(\ell)}\big)^\top\mathbf{W},
\end{gathered}
\end{equation}
where $\ell=1\text{:}S$, and $\mathbf{U}$ is the sum of projection matrices of $\mathbf{K}^{(i)}\mathbf{X}^{(i)}$:
\begin{equation}
\begin{gathered}\label{eq:SumPrj}
\mathbf{U} = \sum_{\ell=1}^{S}\mathbf{I}_{L} - \mathbf{P}^{(\ell)}_{\mathbf{K}\mathbf{X}}= \\ \sum_{\ell=1}^{S}\mathbf{I}_{L} - \mathbf{K}^{(\ell)}\mathbf{X}^{(\ell)}\Big(\big(\mathbf{K}^{(\ell)}\mathbf{X}^{(\ell)}\big)^\top\mathbf{K}^{(\ell)}\mathbf{X}^{(\ell)}+\epsilon\mathbf{I}_{V}\Big)^{-1}\big(\mathbf{K}^{(\ell)}\mathbf{X}^{(\ell)}\big)^\top.
\end{gathered}
\end{equation}
\emph{Proof.} Please see Proofs section.
\end{lemma}
The main difficulty in equation \eqref{eq:GtoA} is calculating the project matrices. Here, this paper uses the approach proposed in \cite{arora12} for determining the projection matrix that can significantly reduce the time complexity. 
\begin{lemma}\label{lm:Project}\emph{For any rectangular real matrix $\mathbf{X}$, a regularized projection can be written as follows:}
\begin{equation}
\begin{gathered}
\mathbf{P} = \mathbf{X}\Big(\mathbf{X}^\top\mathbf{X} + \epsilon\mathbf{I}\Big)\mathbf{X}^\top = \mathbf{A}\mathbf{D}\Big(\mathbf{A}\mathbf{D}\Big)^\top,
\end{gathered}
\end{equation}
where $\epsilon$ is the regularization term, $\mathbf{A}$ denotes the left unitary matrix of $\mathbf{X} \overset{SVD}{=} \mathbf{A}\mathbf{\Sigma}\mathbf{B}^\top$, and $\mathbf{D}$ is a term based on the singular values of $\mathbf{X}$:
\begin{equation}
\begin{gathered}
\mathbf{D}^\top\mathbf{D} = \mathbf{\Sigma}\Big(\mathbf{\Sigma}\mathbf{\Sigma} + \epsilon\mathbf{I}\Big)^{-1}\mathbf{\Sigma}.
\end{gathered}
\end{equation}
\emph{Proof.} We just substituted rank-$m$ SVD of $\mathbf{X}$ in the definition of the projection matrix. Please refer to \cite{arora12} for more information.
\end{lemma}

Lemma \ref{lm:Project} can be employed in equation \eqref{eq:SumPrj} for calculating the projection matrices, where the rank in SVD is equal to the number of stimulus categories (rank-$L$ SVD, $m=L$). By considering equation \eqref{eq:GtoA}, the optimization of SHA can be expressed as follows: \cite{rastogi15}
\begin{equation}\label{eq:MGGL}
\mathbf{U}\mathbf{W} = \mathbf{W}\mathbf{\Lambda},
\end{equation}
where $\Lambda$ and $\mathbf{W}$ respectively denote the eigenvalues and eigenvectors of the matrix $\mathbf{U}$. We can apply Cholesky decomposition \cite{rastogi15} to $\mathbf{U} = \widetilde{\mathbf{U}}\widetilde{\mathbf{U}}^\top$, where $\widetilde{\mathbf{U}}\in \mathbb{R}^{L \times \mu}$, $\mu = L \times S$, is denoted as follows:
\begin{equation}\label{eq:tidleU}
\begin{split}
\widetilde{\mathbf{U}} =  \big[\mathbf{I}_{T} - \mathbf{A}^{(1)}\mathbf{D}^{(1)}\dots\mathbf{I}_{T} - \mathbf{A}^{(S)}\mathbf{D}^{(S)}\big],
\end{split}
\end{equation}
where $\mathbf{W}$ can be considered as the left singular vectors of $\widetilde{\mathbf{U}}=\mathbf{W}\widetilde{\mathbf{\Sigma}}\widetilde{\mathbf{B}}$ \cite{rastogi15}. Since $\widetilde{\mathbf{U}}$ may be too large in order to fit in memory, this paper utilizes Incremental PCA \cite{brand02} for calculating the left singular vectors.

In order to mapped features in both training-set and testing-set, an unsupervised template in the voxel-space is defined as follows by using the supervised shared space:
\begin{equation}\label{eq:W2G}
\begin{gathered}
\mathbf{G} = \frac{1}{S}\Big(\sum_{\ell=1}^{S}\mathbf{W}^T\mathbf{K}^{(\ell)}\Big)^\top,
\end{gathered}
\end{equation}
where $\mathbf{G} \in \mathbb{R}^{T\times V}$ is the unsupervised template. It is worth noting that $\mathbf{W}\text{ and }\mathbf{G}$ are both the shared representational space for neural activities in two different levels, i.e., $\mathbf{W}$ is defined in category-level and $\mathbf{G}$ is denoted in voxel-level.

\subsection{Calculating Mapped Features}
In this section, the mapped features are calculated based on the unsupervised template. While $\mathbf{G}$ is determined in the training-stage by using supervised information, the mapping procedure is completely unsupervised for utilizing in both the training-phase and testing-phase. In the classical Hyperalignment techniques, the mapped features can be calculated by employing following objective function:
\begin{equation}
\begin{split}\label{eq:HAMappings}
\underset{\mathbf{R}^{(i)}}{\min}\Bigg\{\sum_{i=1}^{S}\Big\|\mathbf{X}^{(i)}\mathbf{R}^{(i)} - \mathbf{G}\Big\|^2_F\Bigg\},
\end{split}
\end{equation}
where ${\mathbf{G}}$ is generated by using equation \eqref{eq:W2G} in the training-phase. Then, the mapping matrix can be denoted as follows: \cite{xu12,benton17}
\begin{equation}
\begin{split}\label{eq:SupMappings}
\mathbf{R}^{(\ell)} = \Big(\big(\mathbf{X}^{(\ell)}\big)^\top\mathbf{X}^{(\ell)}+\epsilon\mathbf{I}_{V}\Big)^{-1}\big(\mathbf{X}^{(\ell)}\big)^\top\mathbf{G}.
\end{split}
\end{equation}
However, there is a severe issue in equation \eqref{eq:HAMappings} and equation \eqref{eq:SupMappings} for real-world problems. We need to calculate $\mathbf{R}^{(\ell)} \in \mathbb{R}^{V\times V}$ for $\ell=1\text{:}S$, where the number of voxels ($V$) is almost big and thus mapping matrix $\mathbf{R}^{(\ell)} \in \mathbb{R}^{V \times V}$ needs a lot of memory, e.g., $\mathbf{R}^{(\ell)}$ is around $300GB$ for whole-brain fMRI datasets in the MNI space. In order to deal with this issue, we directly calculate the mapped features without calculating $\mathbf{R}^{(\ell)}$. We first denote the mapped features as follows and then substitute equation \eqref{eq:SupMappings} on the definition:
\begin{equation}
\begin{gathered}\label{eq:Mappings}
\mathbf{Z}^{(\ell)} = \mathbf{X}^{(\ell)}\mathbf{R}^{(\ell)} = \\ \mathbf{X}^{(\ell)}\Big(\big(\mathbf{X}^{(\ell)}\big)^\top\mathbf{X}^{(\ell)}+\epsilon\mathbf{I}_{V}\Big)^{-1}\big(\mathbf{X}^{(\ell)}\big)^\top\mathbf{G} = \mathbf{P}^{(\ell)}_{\mathbf{X}}\mathbf{G},
\end{gathered}
\end{equation}
where $\mathbf{P}^{(\ell)}_{\mathbf{X}}$ is the projection of the neural activities $\mathbf{X}^{(\ell)}$ that can be calculated by using Lemma \ref{lm:Project}. Further, $\mathbf{Z}^{(\ell)} \in \mathbb{R}^{T \times V}$ is the mapped features in the shared space. Algorithm \ref{alg:MVP} illustrates a general framework for fMRI analysis by using SHA method. Since there is no iteration in this algorithm, it provides acceptable runtime for real-world problems.

\begin{algorithm}[!t]
	\caption{Supervised Hyperalignment (SHA)}
	\label{alg:MVP}
	\begin{algorithmic}
		\STATE {\bfseries Input:} Train Set $\mathbf{X}^{(i)}, i=1\text{:}S$, Test Set $\mathbf{\bar{X}}^{(j)}, j=1\text{:}\bar{S}$,\\
		\quad Training Class Labels $\mathbf{Y}^{(i)}$, Regularization $\epsilon=10^{-4}$.
		\STATE {\bfseries Output:} Classification Performance ($ACC$, $AUC$).\\
		\STATE {\bfseries Method:}\\
		\quad01. Initialize $\mathbf{K}^{(\ell)}\text{, }\mathbf{H}$.\\
		\quad02. Calculate $\widetilde{\mathbf{U}}$ based on equation \eqref{eq:tidleU}.\\
		\quad03. Apply SVD \cite{brand02} on  $\widetilde{\mathbf{U}}$ for decomposing $\mathbf{W}$. \\
		\quad04. Calculate $\mathbf{G}$ by using $\mathbf{W}$ and equation \eqref{eq:W2G}.\\
		\quad05. Generate $\mathbf{Z}^{(\ell)}, \ell=1\text{:}S$ as the training-set.\\
		\quad06. Generate $\mathbf{\bar{Z}}^{(\ell)}, \ell=1\text{:}\bar{S}$ as the testing-set.\\
		\quad07. Train a classifier by $\mathbf{Z}^{(\ell)}, \mathbf{Y}^{(\ell)} \text{ for }\ell=1\text{:}S$.\\
		\quad08. Evaluate the classifier by using $\mathbf{\bar{Z}}^{(\ell)}, \ell=1\text{:}\bar{S}$.
	\end{algorithmic}
\end{algorithm}

\section{Experiments}
To evaluate the performance of the proposed SHA, we have compared it with 10 different existing methods: \emph { raw data without functional alignment (NONE);  the original Hyperalignment (HA) \cite{haxby11};  Regularized Hyperalignment (RHA) \cite{xu12}; Kernel-based Hyperalignment (KHA) \cite{lorbert12}; SearchLight Hyperalignment (SLHA) \cite{guntupalli16}; Joint SVD Hyperalignment (SVDHA) \cite{chen14};  Shared Response Model (SRM) \cite{chen15}; Deep Hyperalignment (DHA) \cite{Tony17NIPS}; Semi-Supervised SRM (SS-SRM) \cite{turek16}; and Local Discriminant Hyperalignment (LDHA) \cite{tony17a}}. Further, we evaluate the performance of SHA approach by using two optimization approaches, i.e., \emph{SHA-R} that uses the regularization and optimization technique in \cite{xu12}, and \emph{SHA}, which utilizes the proposed optimization in this paper.

The experimental evaluation has been focused on 
\begin{enumerate}
\item performance of correlation analysis of different HA methods in the training phase, i.e., how each method can maximize the correlation within-class stimuli and minimize the between-class stimuli; 
\item performance of different HA methods in the post-alignment classification using datasets of simple tasks and complex tasks respectively;
\item computing performance/runtime analysis.
\end{enumerate}

With regard to implementation, we have used Generalized CCA proposed in \cite{rastogi15} in order to generate the original HA algorithm. Moreover, regularized parameters ($\alpha,\beta$) are optimized for RHA based on \cite{xu12}, where these parameters generate the lowest alignment error on the training-phase. In addition, Gaussian kernel that is reported as the best kernel in the original paper is employed for generating the KHA results \cite{lorbert12}. Like the original paper \cite{Tony17NIPS}, we consider $3$ hidden layer ($C=5$) for DHA, the number of units in the intermediate layers are considered $L\times V$, and deep network is trained by using $\eta=10^{-4}$ learning rate.  
For SS-SRM, we also considered $\gamma=1.0$ and $\alpha=0.5$ \cite{turek16}. It is worth noting that this paper employs BrainIAK library\footnote{Available at https://brainiak.org} for running SRM and SS-SRM. The number of features in all of the alignment techniques are considered $\min(V, T)$. Further, the number of iterations and regularization term for all of the mentioned methods are respectively considered as $\tau=10$ and $\epsilon=10^{-4}$. These algorithms are implemented in our GUI-based toolbox, called easy fMRI\footnote{Easy fMRI available at https://easyfmri.gitlab.io/ \\Supervised hyperalignmet code available at: https://gitlab.com/easyfmri/easyfmri/blob/master/Hyperalignment/SHA.py}, and run on a computer with certain specifications\footnote{Main:~Giga X399, CPU:~AMD Ryzen Threadripper 2920X~(24$\times$3.5~GHz), RAM:~64GB, GPU:~NVIDIA GeForce RTX 2080 SUPER~(8~GB memory), OS:~Fedora~31, Python:~3.7.5, Pip:~19.3.1, Numpy:~1.16.5, Scipy:~1.2.1, Scikit-Learn:~0.21.3., MPI4py:~3.0.1, PyTorch:~1.2.0} in order to generate experiments.

\begin{table*}[t]
	\caption{The datasets}
	\vskip -0.1in
	\label{tbl:Datasets}
	\begin{small}
		\begin{center}	
			\rowcolors{1}{light-gray}{white}
			\begin{tabular}{ccccccccccccc}
				\hline
				ID       & X   & Y   & Z  & S  & R   & L  & T   (rest)& V     & FWHM & TR  & TE & Scanner  \\
				\hline
				DS005    & 53  & 63  & 52 & 16 & 48  & 2  & 240 (153) & 450   & 5 mm & 2   & 30 & S 3T   \\
				DS105    & 79  & 95  & 79 & 6  & 71  & 8  & 121 (37)  & 1963  & 5 mm & 2.5 & 30 & GE 3T \\
				DS107    & 53  & 63  & 52 & 49 & 98  & 4  & 164 (42)  & 932   & 6 mm & 2   & 28 & S 3T  \\
				DS113    & 160 & 160 & 36 & 20 & 160 & 10 & 451 (72)  & 2400  & 5 mm & 2.3 & 22 & S 7T  \\
				DS116    & 53  & 63  & 40 & 17 & 102 & 2  & 170 (64)  & 2532  & 5 mm & 2   & 25 & P 3T  \\
				DS117    & 64  & 61  & 33 & 19 & 171 & 2  & 210 (76)  & 524   & 5 mm & 2   & 30 & S 3T  \\
				Raiders  & 78  & 78  & 54 & 10 & 10  & 7  & 924 (142) & 980   & 4 mm & 3   & 30 & S 3T\\
				CMU      & 51  & 61  & 23 & 9  & 9   & 12 & 402 (42)  & 17326 & N/A & 1   & 30 & S 3T\\
				\hline
			\end{tabular}
		\end{center}
		X, Y, Z denote the size of 3D images; S is the number of subjects; R denotes the number of all runs; L is the number of classes; T denotes the number of time points; rest is the number of time points without label after temporal-alignment; V denotes the number of voxels in ROI; FWHM is Full Width at Half Maximum; TR denotes Time of Repetition in Second; TE is Echo Time in millisecond; Scanners are S (Siemens), GE (General Electric), P (Philips) with two magnet powers, i.e., 3 or 7 Tesla. 
	\end{small}
\end{table*}
\vskip -0.1in

\subsection{Datasets}
This paper utilizes 8 datasets, mostly shared by Open~Neuro\footnote{Available at https://openneuro.org/} for empirical studies. We divided it into two categories: simple and complex cognitive tasks. \textbf{\emph{Simple cognitive task}} refers to a simple cognitive task where only visual or audio stimuli involved such as watching grayscale photos or taping keys, while a \textbf{\emph{complex cognitive task}} refers to a complex task such as watching a movie scene with different visual and audio stimuli. The datasets related to simple cognitive tasks are DS005, DS105, DS107, DS116, DS117, and CMU. The datasets related to complex cognitive tasks are Raiders and DS113.

Table \ref{tbl:Datasets} depicts list of these datasets. In this table, DS005 includes fMRI images related to $2$ classes of risk tasks with $50/50$ chance of selection for $48$ subjects. The Region of Interest (ROI) for this dataset is selected by using the original paper \cite{tom07}. The next dataset is DS105 that includes $8$ classes (categories) of visual stimuli for $6$ subjects, i.e., gray-scale images of houses, faces, scissors, cats, shoes, bottles, chairs, and nonsense patterns (scrambles). As the ROI in this dataset, the neural activities in the Ventral Temporal (VT) cortex are analyzed. Please see \cite{haxby11,haxby14} for technical information. The third dataset is DS107 that includes fMRI images for $98$ subjects. This dataset contains $4$ classes of visual stimuli, i.e., photos of consonants, objects, words, and scrambles (nonsense). Here, ROI is defined based on \cite{duncan09}. DS113 is the fourth dataset that is collected from 20 subjects, who watched `Forrest Gump (1994)' movie as the visual stimuli. Please refer to \cite{hanke14} for technical information. The next dataset is DS116, which contains EEG signals and fMRI images. We only employ the fMRI images in this section. This dataset includes audio and visual stimuli of oddball tasks, and the ROI is denoted by using the original paper \cite{walz13}. The sixth dataset is DS117, which contains MEG and fMRI images. This section employs the fMRI images in this dataset. Further, it includes 2 classes of visual stimuli, i.e., photos of human faces, and scrambles. The neural responses in the VT cortex are used to generate MVP models. Please see \cite{wakeman15} for more information. As the seventh dataset, Raiders includes fMRI images that are collected from 10 subjects, who watched `Raiders of the Lost Ark (1981)' movie. In this dataset, the VT cortex is selected as the ROI. Please refer to  \cite{chen14,chen15,lorbert12,sabuncu10} for technical information. CMU is the last dataset that includes the neural responses for 9 subjects. It includes 60 different word pictures as the visual stimuli, where they are clustered to 12 semantic categories. Furthermore, the intersection of coordinates is defined across subjects as the ROI in this dataset. Please see \cite{mitchell08,song16} for more information.

All of the mentioned datasets are independently preprocessed by easy fMRI and FSL 6.0.0\footnote{Available at https://fsl.fmrib.ox.ac.uk}, i.e., slice timing, anatomical alignment, normalization, smoothing. The preprocessing information is presented in Table~\ref{tbl:Datasets}. The fMRI images are registered to MNI standard space by using FMRIB's Linear Image Registration Tool (FLIRT) \cite{jenkinson01}, and then the motion correction is applied by utilizing Motion Correction FLIRT (MCFLIRT) \cite{jenkinson02}. The brain area also extracted from fMRI images by using the Brain Extraction (BET) algorithm \cite{smith02}. Further, FMRIB's Improved Linear Model (FILM) method with local autocorrelation correction is utilized for time-series statistical analysis \cite{woolrich01}. We have also provided a preprocessed version of these datasets in MATLAB format\footnote{Available at https://easydata.gitlab.io/}.
\begin{figure*}[!t]
	\begin{center}
		\begin{minipage}{0.98\linewidth}\includegraphics[width=0.98\textwidth]{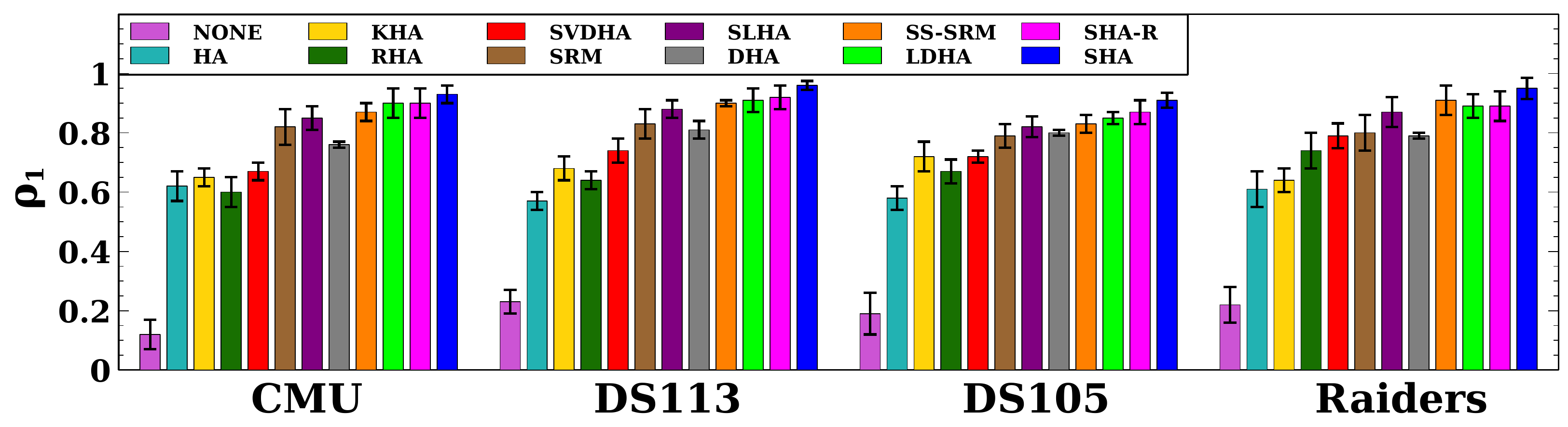}
			\centering {(a) Whole of mapped time series.}
		\end{minipage}
		\begin{minipage}{0.98\linewidth}\includegraphics[width=0.98\textwidth]{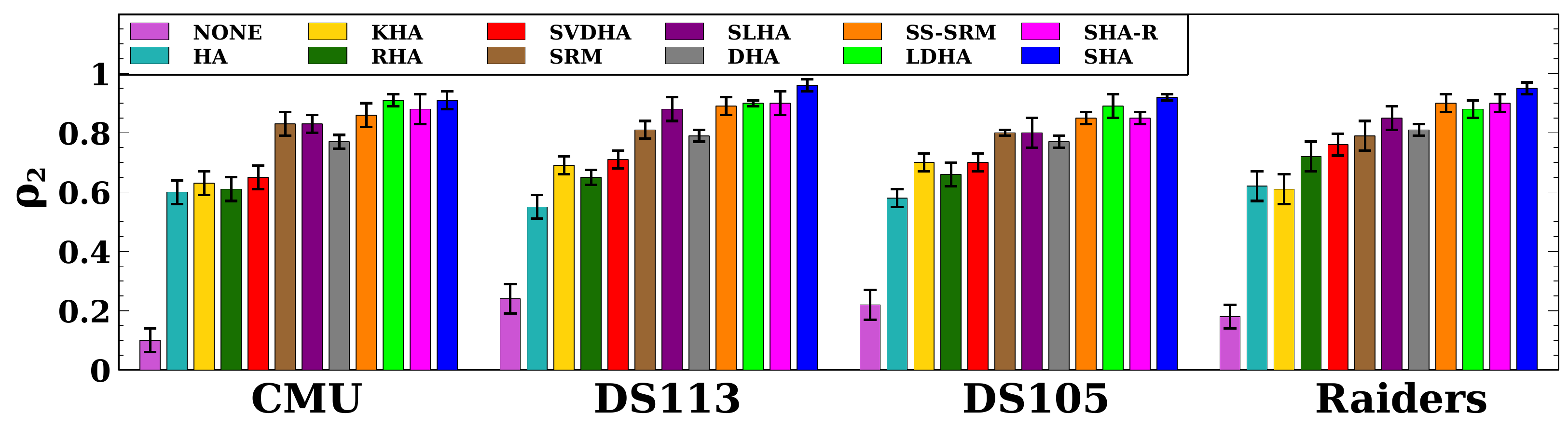}
			\centering {(b) Stimuli in the same category and the same  location.}
		\end{minipage}
		\begin{minipage}{0.98\linewidth}\includegraphics[width=0.98\textwidth]{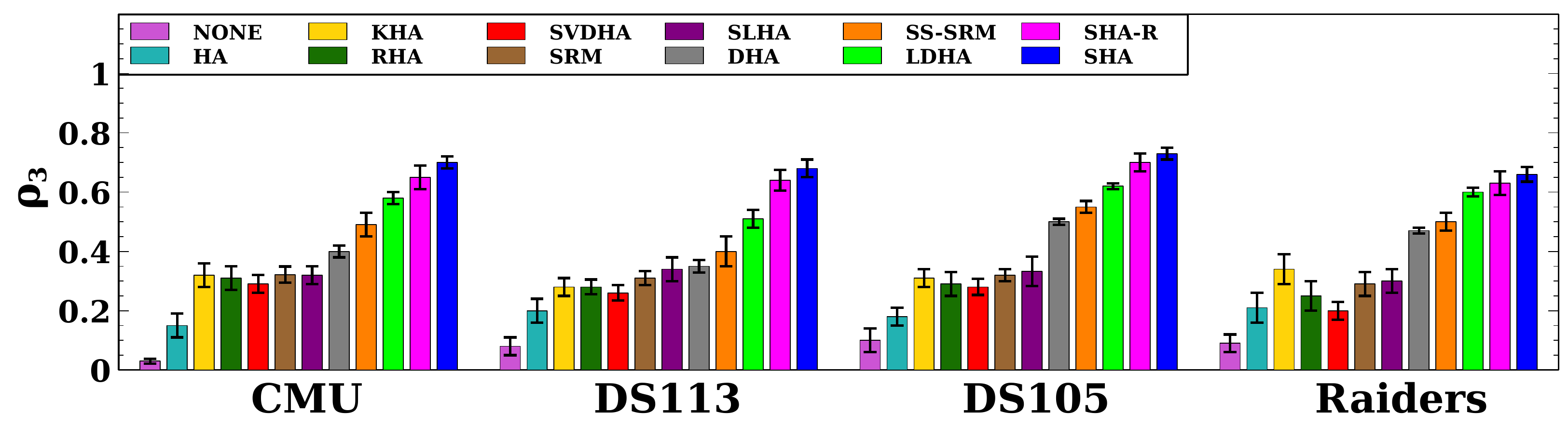}
			\centering {(c) Stimuli in the same category in different locations.}
		\end{minipage}
		\begin{minipage}{0.98\linewidth}\includegraphics[width=0.98\textwidth]{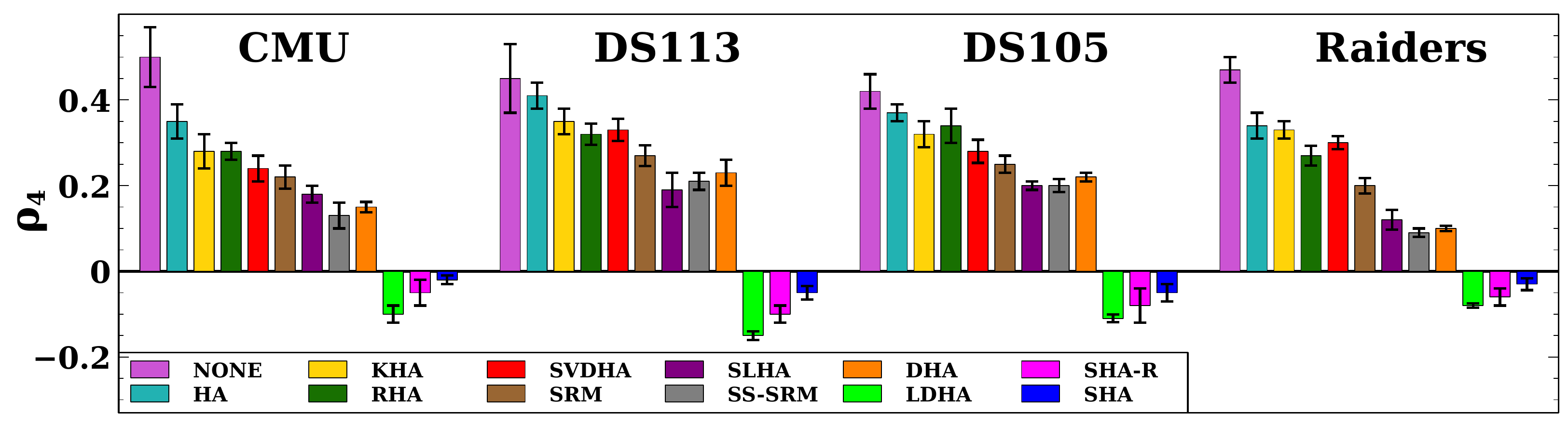}
			\centering {(d) Stimuli in different categories.}
		\end{minipage}
		\caption{Correlation analysis of the mapped neural activities in the training phase across subjects (mean$\pm$std).}\label{fig:Correlation}
	\end{center}
	\vskip -0.2in
\end{figure*}

\begin{figure*}[!t]
	\begin{center}
		\begin{minipage}{\textwidth}
			\includegraphics[width=0.98\textwidth]{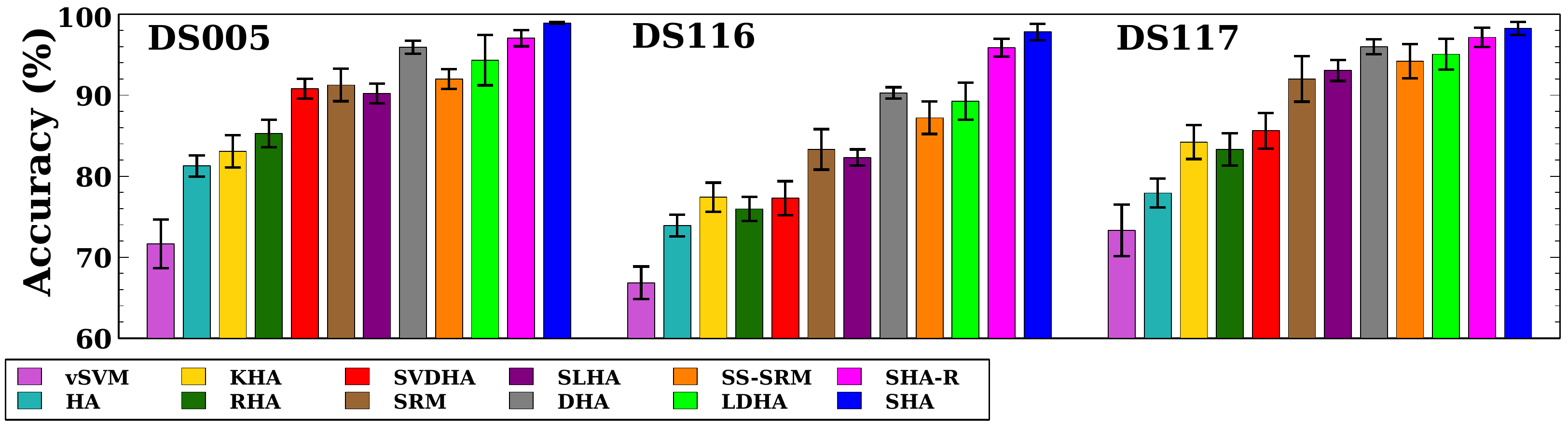}\\ 
			\centering {\small (a) Accuracy of different HA methods in post-alignment \emph{binary} classification }
		\end{minipage}
		\begin{minipage}{\textwidth}
			\includegraphics[width=0.98\textwidth]{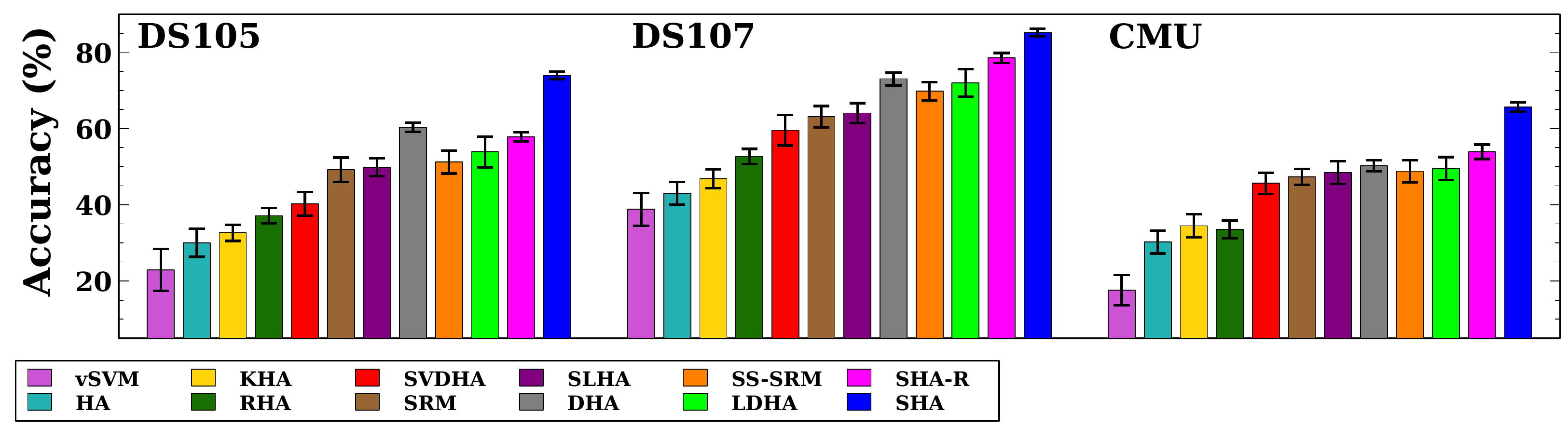}\\
			\centering {\small (b) Accuracy of different HA methods in post-alignment \emph{multi-class} classification}
		\end{minipage}
		\caption{Classification analysis on simple task datasets}
		\label{fig:ClassificationAccuracy}
	\end{center}
	\vskip -0.2in
\end{figure*}

\subsection{Correlation Analysis}
It is a common ground that the better HA methods must provide mappings with highly correlated within-class stimuli and minimal correlation among between-class stimuli \cite{lorbert12,chen15}. To validate the performance of our proposed algorithm, we have conducted correlation analysis of mapped neural activities across subjects in four different levels, i.e., \emph {whole of mapped time series, the stimuli that belong to the same category and the same location in the time series, the stimuli that belong to the same category and different locations in the time series, and the stimuli that belong to the distinctive categories}. Since these analyses are based on stimulus categories, this paper employed datasets in Table \ref{tbl:Datasets} with the largest number of categories, i.e., CMU, DS113, DS105, and Raiders.  

\subsubsection{Correlation analysis between whole of mapped stimuli}

Different HA methods are employed in order to map neural activities to a shared space. Then, the average ($\rho_1$) of Pearson correlations (\emph{corr()}) for every pair of subjects is calculated by using the whole of mapped features $\bigg(\rho_1 = \frac{1}{\Psi_1} \sum_{i=1}^{S}\sum_{j = i+1}^{S}\text{corr}\Big(\mathbf{X}^{(i)}\mathbf{R}^{(i)}, \mathbf{X}^{(j)}\mathbf{R}^{(j)}\Big)\bigg)$, where $\Psi_1=\sfrac{S(S-1)}{2}$. In order to make sense of this experiment, just consider the example of Figure \ref{fig:DifferentHAs}, where the whole of time series include following stimuli $[\mathbf{H1}, \mathbf{B1}, \mathbf{H2}, \mathbf{B2}]$ for two subjects ($\mathbf{S1, S2}$). Based on the setup of the first experiment, we have $\rho_1=\text{corr}\Big(\mathbf{S1\text{:}[\dots]}\text{, }\mathbf{S2\text{:}[\dots]}\Big)$, where $\mathbf{S1\text{:}[\dots]}$ denotes mapped features in the whole of time series belong to the first subject.

Figure \ref{fig:Correlation}.a illustrates the first analysis, where `NONE' denotes the correlation of the preprocessed (i.e., anatomical alignment) datasets without functional alignment ($\mathbf{R}^{(\ell)}=\mathbf{I}_V$). As depicted in this figure, the original neural activities in different brains (subjects) have low correlation in comparison with each other. In fact, this is the main reason in order to use HA methods \cite{haxby11}. Further, this figure shows that different HA methods provide distinctive performances.

\subsubsection{Correlation analysis for stimuli belonging to the same category in the same location of time series}

The average of correlations for each pair of subjects is generated by employing the stimuli that belong to the same category and the same location in the time series. In order to express the setup of this experiment, $L_m$ denotes the number of stimuli in $m\text{-}th$ category. Further, $\mathbf{X}^{(i)}_{[m;n]}$ and $\mathbf{R}^{(i)}_{[m;n]}$ are respectively defined as the neural activities and HA mapping that belong to $i\text{-}th$ subject, $m\text{-}th$ category, and $n\text{-}th$ stimulus. Based on these definitions, the average of correlations in this experiment is calculated as follows:
\begin{equation}
\begin{split}
\rho_2 = \frac{1}{\Psi_2} \sum_{\substack{i=1\\j = i+1}}^{S}\sum_{m=1}^{L}\sum_{n = 1}^{L_m}
\text{corr}\Big(\mathbf{X}^{(i)}_{[m;n]}\mathbf{R}^{(i)}_{[m;n]}\text{, }\mathbf{X}^{(j)}_{[m;n]}\mathbf{R}^{(j)}_{[m;n]}\Big),
\end{split}
\end{equation}
where $\Psi_2=\frac{1}{2}S(S-1)\big(\sum_{m=1}^{L}L_m\big)$. Based on the example of Figure \ref{fig:DifferentHAs}, we also have following comparisons:

\begin{equation*}
\begin{split}
\rho_2 = \frac{1}{4}
\bigg(\text{corr}\Big(\mathbf{S1\text{:}}[\mathbf{H1}]\text{, }\mathbf{S2\text{:}}[\mathbf{H1}]\Big)+
\text{corr}\Big(\mathbf{S1\text{:}}[\mathbf{B1}]\text{, }\mathbf{S2\text{:}}[\mathbf{B1}]\Big)\\
+\text{corr}\Big(\mathbf{S1\text{:}}[\mathbf{H2}]\text{, }\mathbf{S2\text{:}}[\mathbf{H2}]\Big)+
\text{corr}\Big(\mathbf{S1\text{:}}[\mathbf{B2}]\text{, }\mathbf{S2\text{:}}[\mathbf{B2}]\Big)\bigg),
\end{split}
\end{equation*}
where $\mathbf{S1\text{:}}[\mathbf{H1}]$ is the mapped neural activities that belong to the first subject and the first house stimulus ($\mathbf{H1}$). Figure \ref{fig:Correlation}.b depicts the results of the second experiment. These results are similar to the first experiment. There are two reasons for this similarity. First of all, neural activities in training-set are time synchronized. Secondly, most of the HA methods used injective mappings in order to align these time points to a shared space \cite{xu12,chen15,haxby11}.

\begin{figure*}[!h]
	\begin{center}
		\begin{minipage}{0.24\linewidth}
			\includegraphics[width=0.98\textwidth,height=0.6\linewidth]{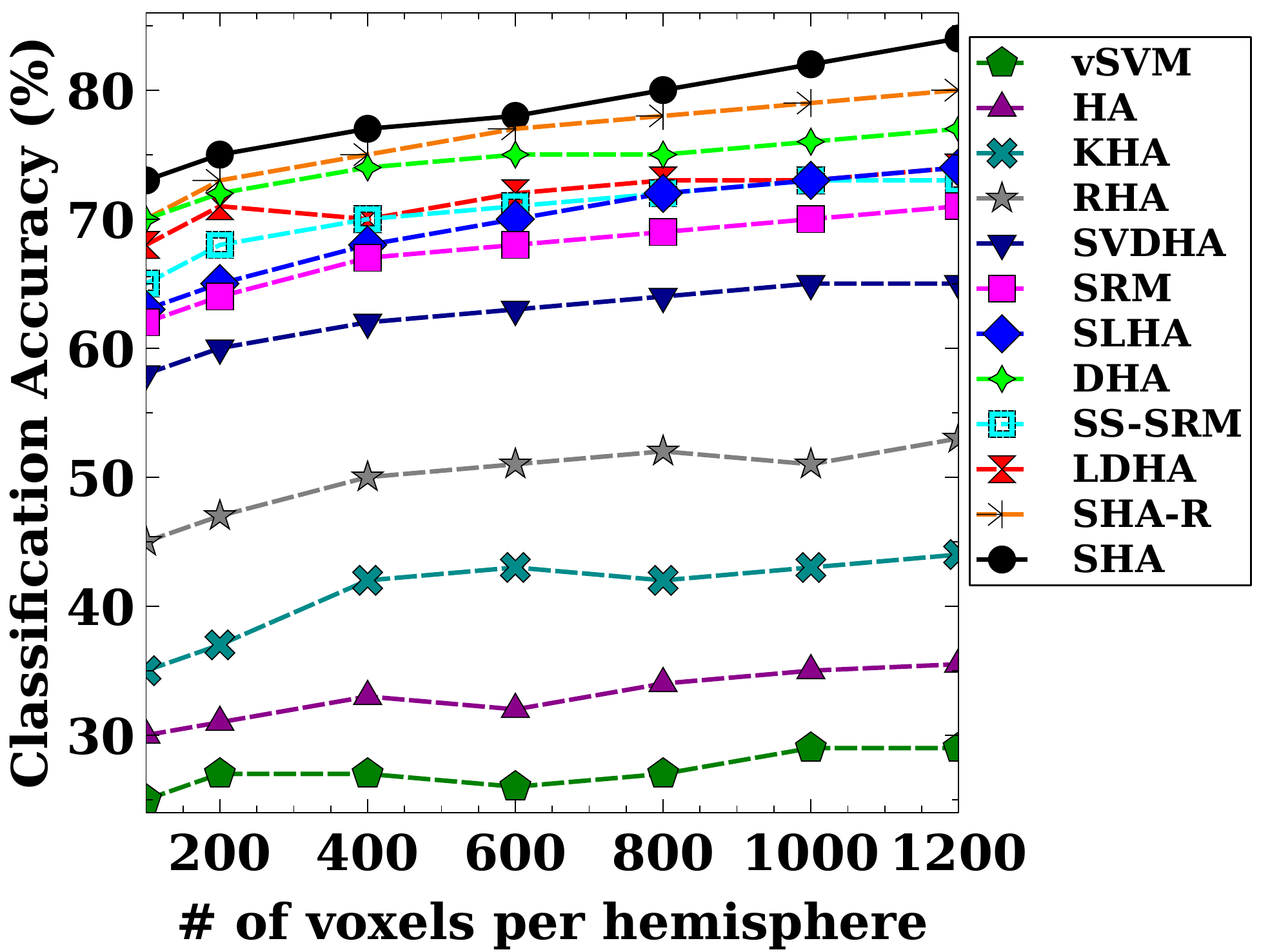}\\
			\centering {\small (a) DS113 (TRs = 100)}
		\end{minipage}
		\begin{minipage}{0.24\linewidth}
			\includegraphics[width=0.98\textwidth,height=0.6\linewidth]{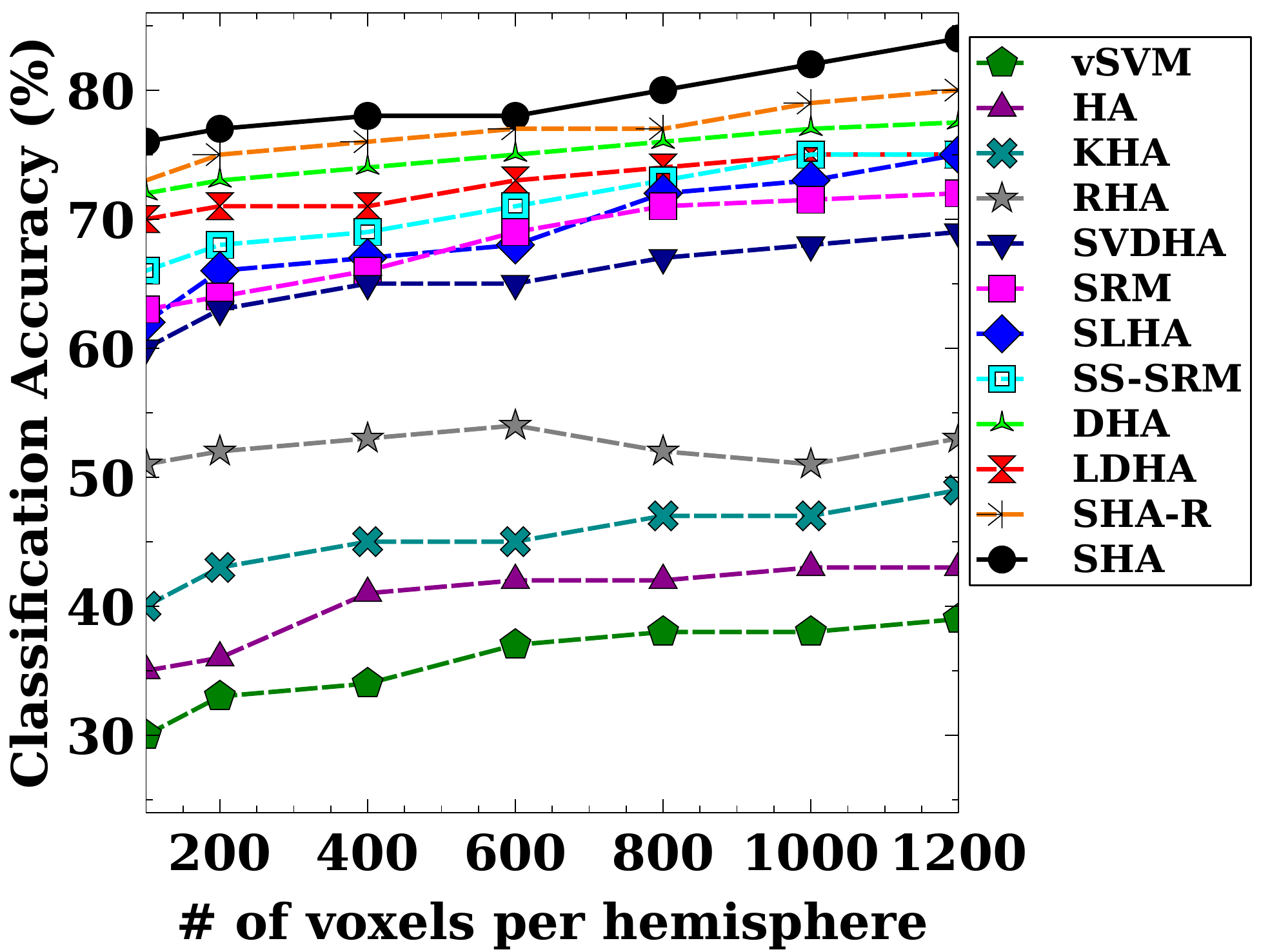}\\
			\centering {\small (b) DS113 (TRs = 400)}
		\end{minipage}
		\begin{minipage}{0.24\linewidth}
			\includegraphics[width=0.98\textwidth,height=0.6\linewidth]{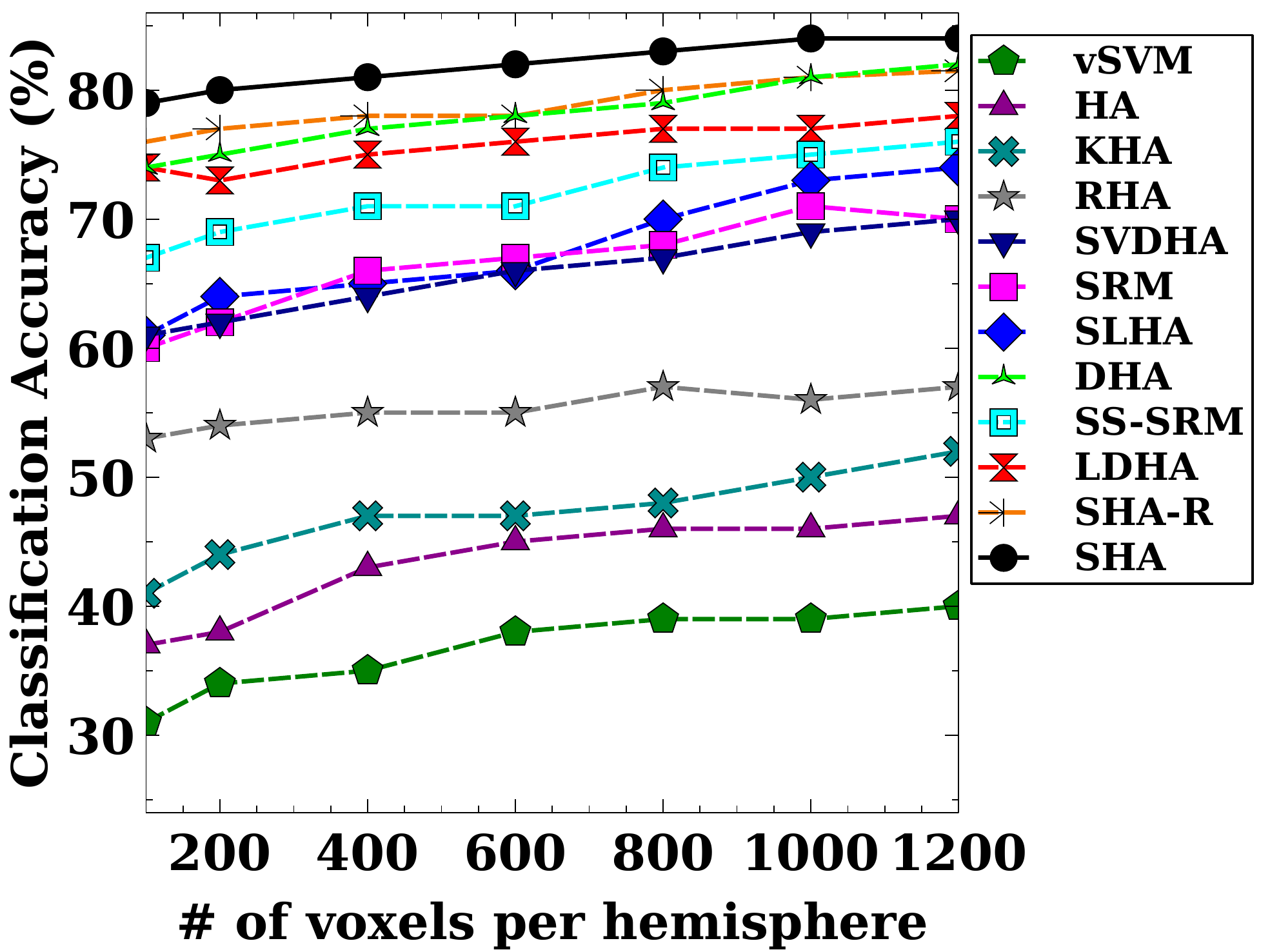}\\
			\centering {\small (c) DS113 (TRs = 800)}
		\end{minipage}
		\begin{minipage}{0.24\linewidth}
			\includegraphics[width=0.98\textwidth,height=0.65\linewidth]{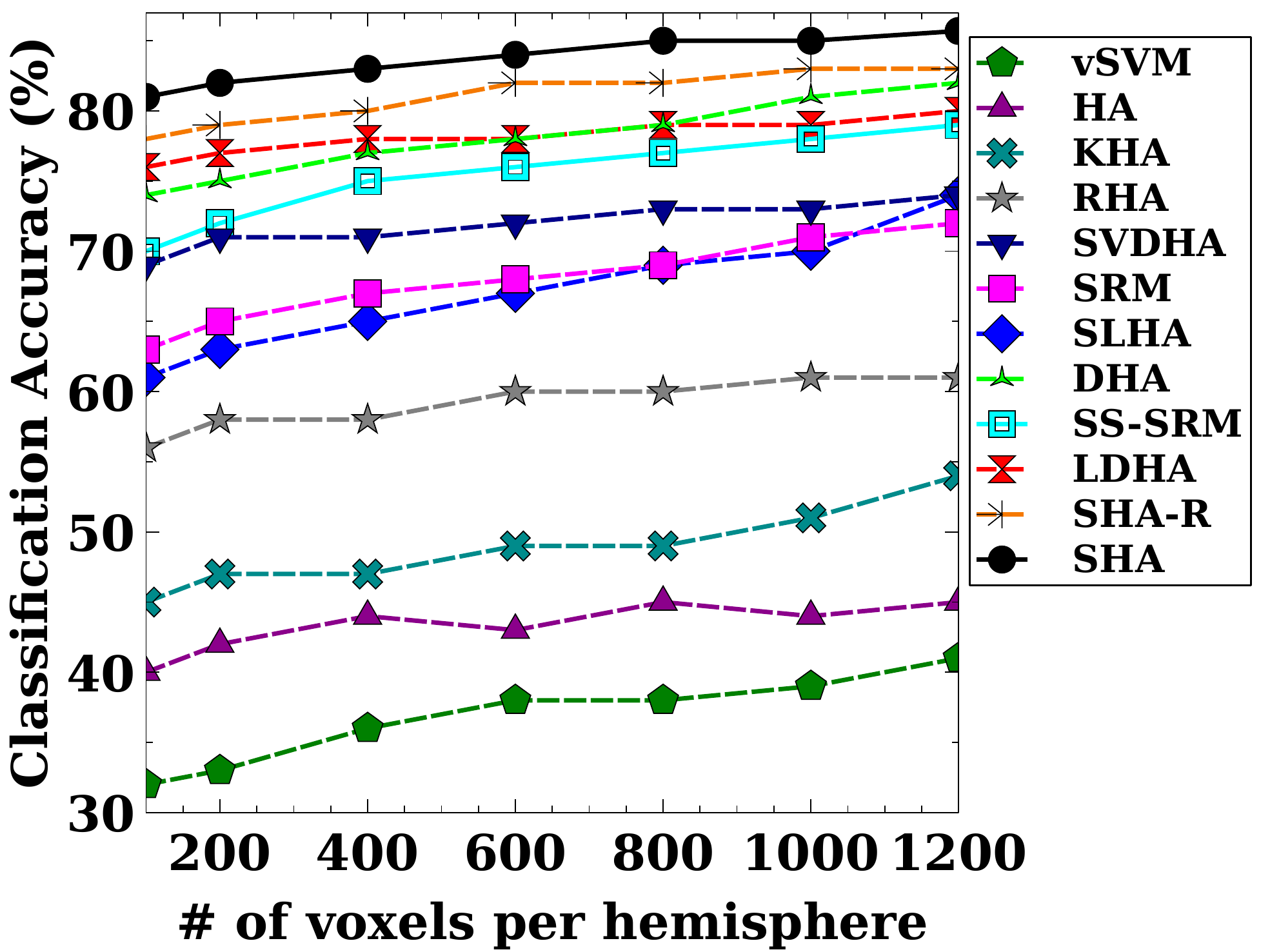}\\
			\centering {\small (d) DS113 (TRs = 2000)}
		\end{minipage}
		\begin{minipage}{0.24\linewidth}
			\includegraphics[width=0.98\textwidth,height=0.6\linewidth]{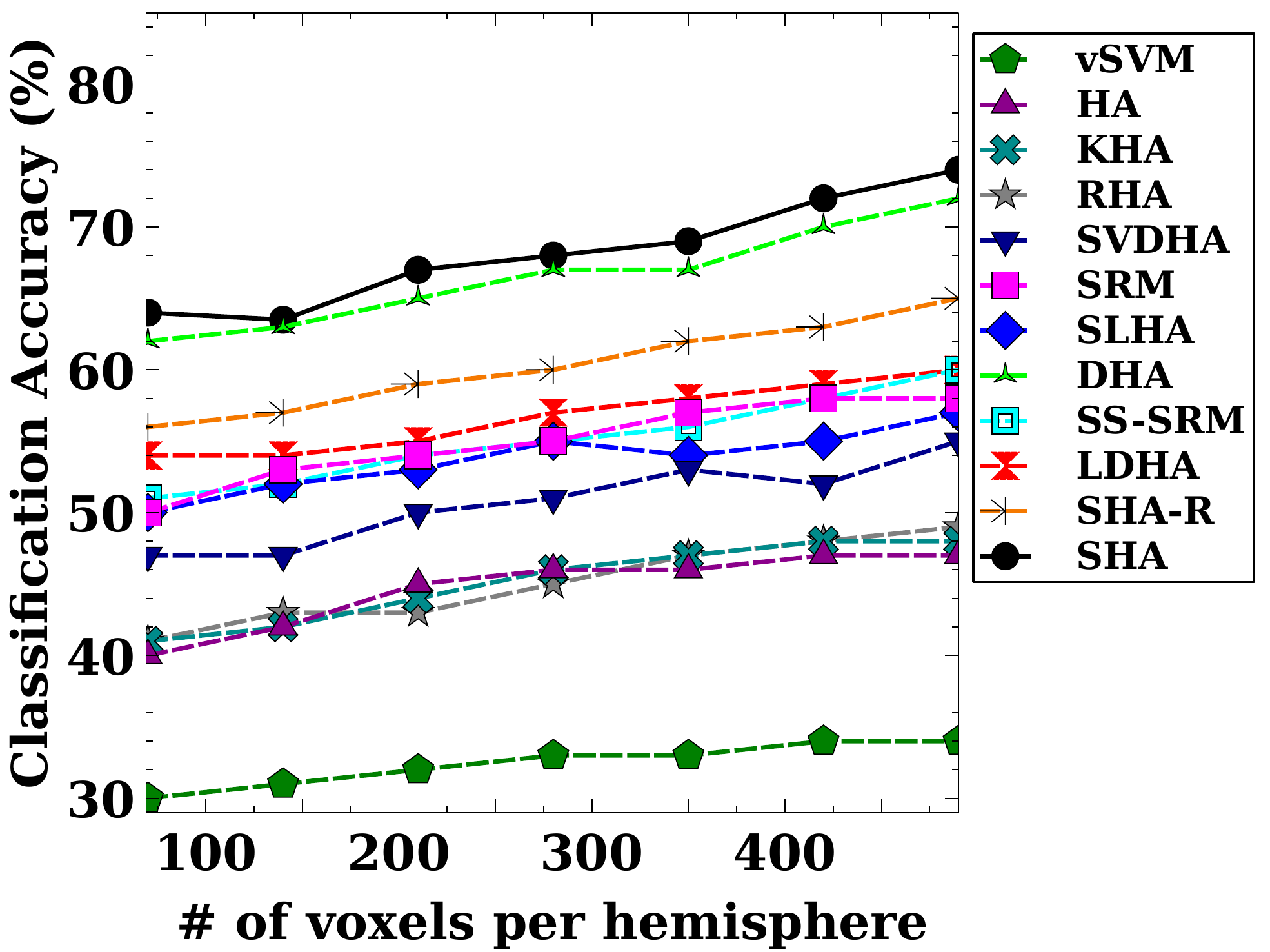}\\
			\centering {\small  (e) Raiders (TRs = 100)}
		\end{minipage}	
		\begin{minipage}{0.24\linewidth}
			\includegraphics[width=0.98\textwidth,height=0.6\linewidth]{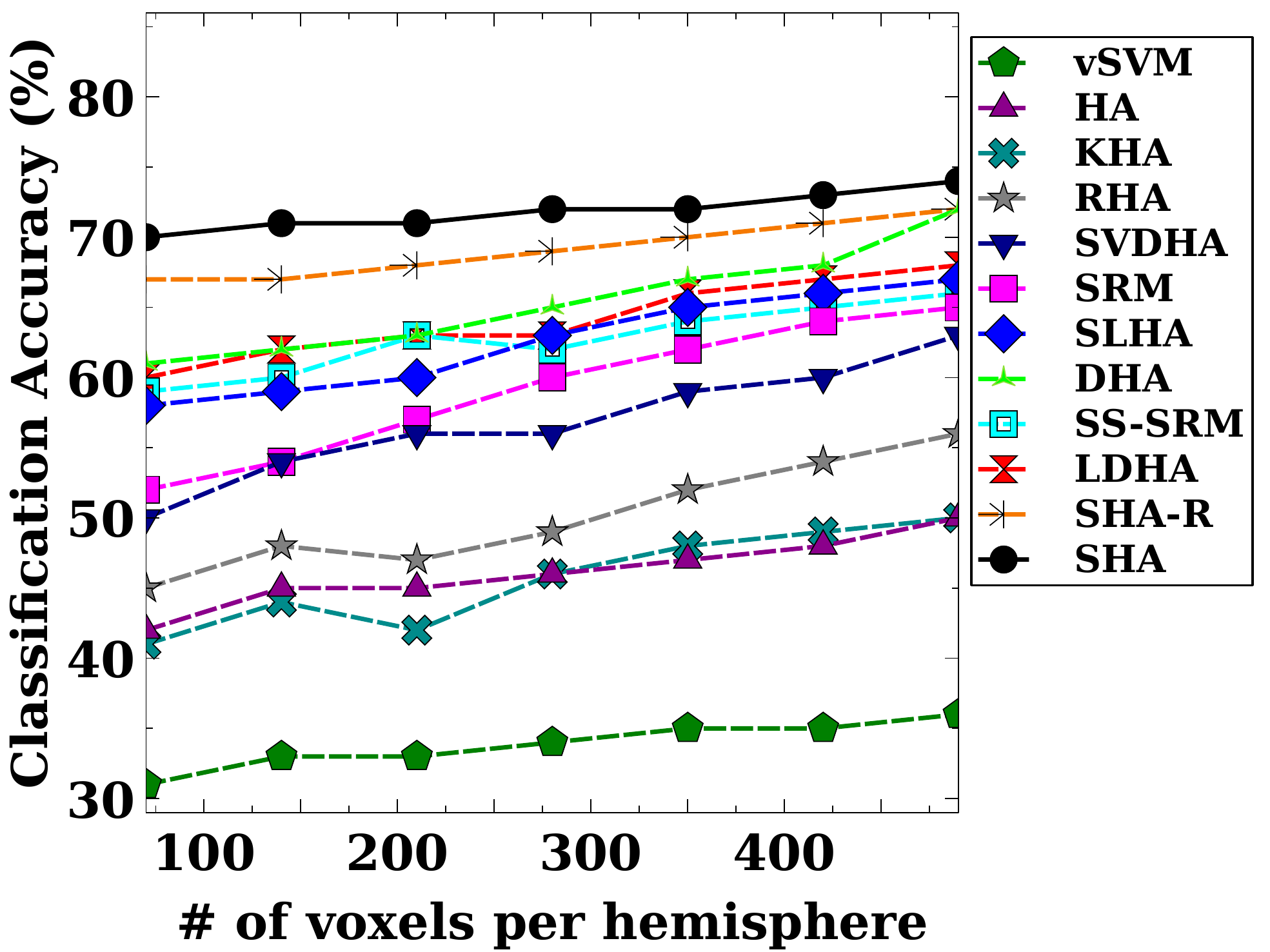}\\
			\centering {\small (f) Raiders (TRs = 400)}
		\end{minipage}				
		\begin{minipage}{0.24\linewidth}
			\includegraphics[width=0.98\textwidth,height=0.6\linewidth]{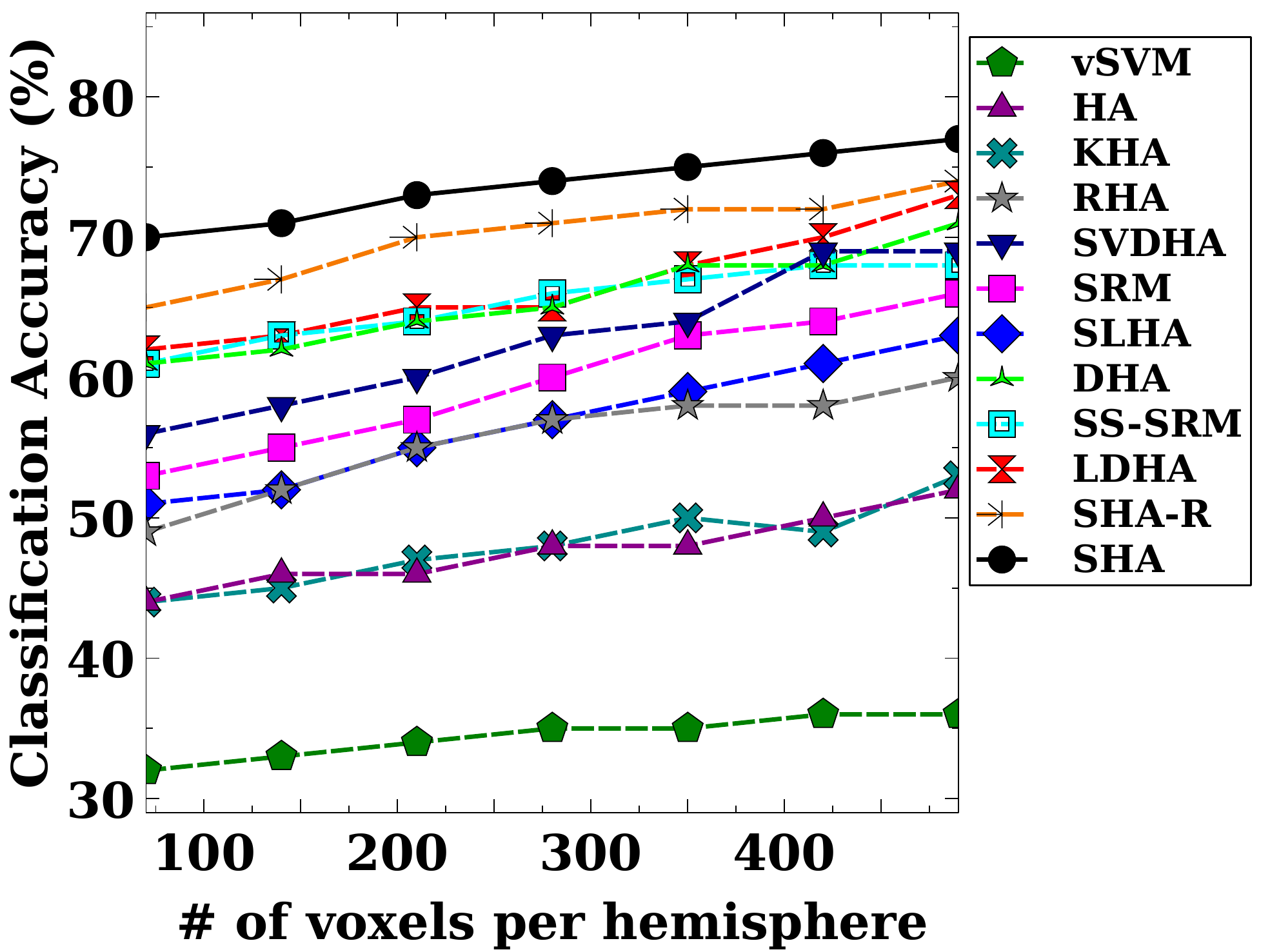}\\
			\centering {\small  (g) Raiders (TRs = 800)}
		\end{minipage}			
		\begin{minipage}{0.24\linewidth}
			\includegraphics[width=0.98\textwidth,height=0.6\linewidth]{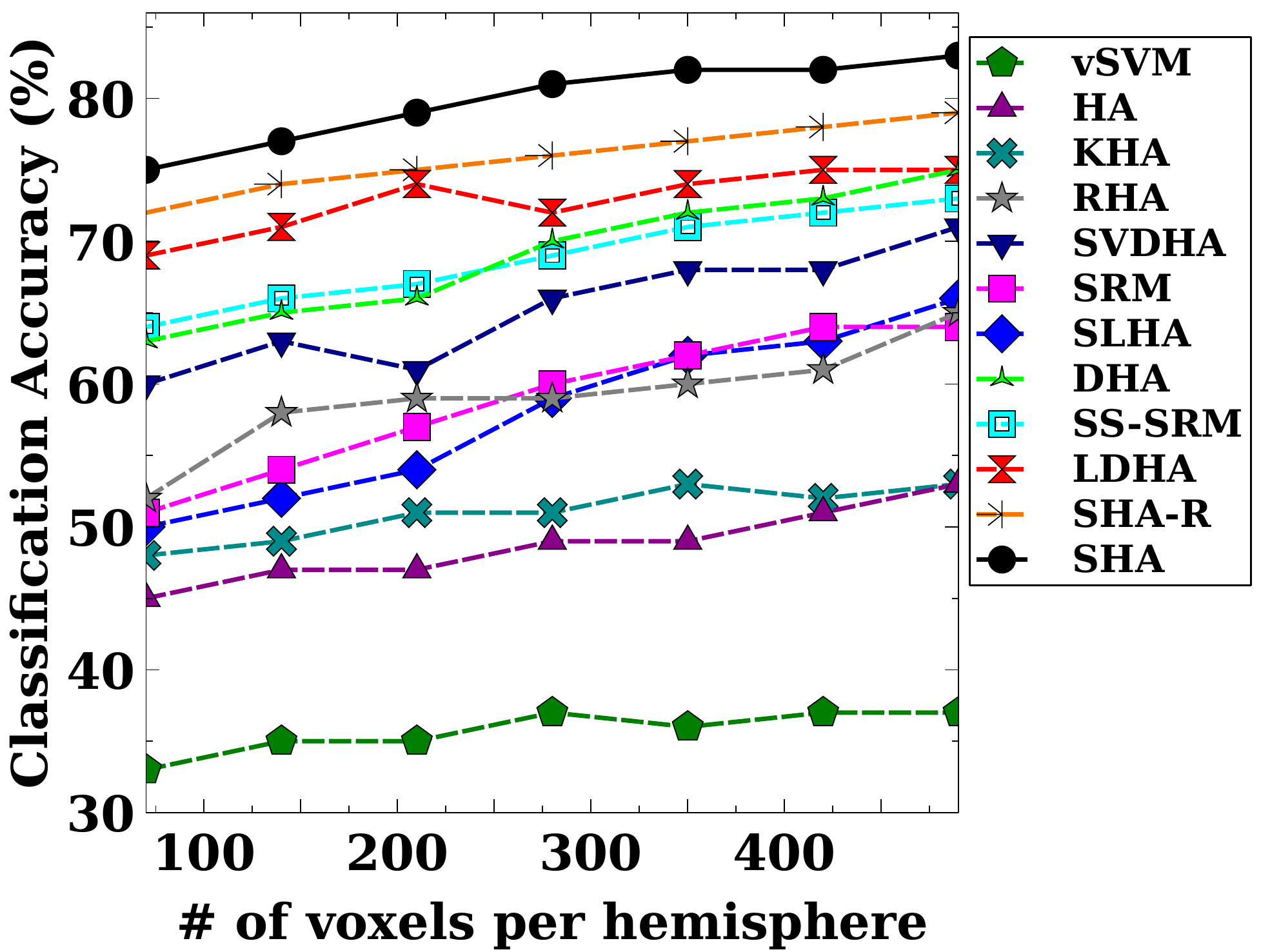}\\
			\centering {\small (h) Raiders (TRs = 2000)}
		\end{minipage}			
		\caption{Comparison of different HA algorithms on complex task datasets}
		\label{fig:Movies}
	\end{center}
	\vskip -0.2in
\end{figure*}

\subsubsection{Correlation analysis for stimuli belonging to the same category in the different location of time series}

The average of correlations for every pair of subjects is calculated by using the stimuli that belong to the same category and the different locations of the time series. Average of correlations in this experiment is estimated as follows:
\begin{equation}
\begin{split}
\rho_3 = \frac{1}{\Psi_3} \sum_{\substack{i=1\\j = i+1}}^{S}\sum_{m=1}^{L}\sum_{\substack{n,\ell = 1\\n\neq\ell}}^{L_m}
\text{corr}\Big(\mathbf{X}^{(i)}_{[m;n]}\mathbf{R}^{(i)}_{[m;n]}\text{, }\mathbf{X}^{(j)}_{[m;\ell]}\mathbf{R}^{(j)}_{[m;\ell]}\Big),
\end{split}
\end{equation}
where $\Psi_3=\frac{1}{2}S(S-1)\Big(\sum_{m=1}^{L}\big(L_m^2 - L_m\big)\Big)$. By considering the example of Figure \ref{fig:DifferentHAs}, we also have following comparisons:
\begin{equation*}
\begin{gathered}
\rho_3 = \frac{1}{4}\bigg(
\text{corr}\Big(\mathbf{S1\text{:}}[\mathbf{H1}]\text{, }\mathbf{S2\text{:}}[\mathbf{H2}]\Big)+
\text{corr}\Big(\mathbf{S1\text{:}}[\mathbf{B1}]\text{, }\mathbf{S2\text{:}}[\mathbf{B2}]\Big)\\+
\text{corr}\Big(\mathbf{S1\text{:}}[\mathbf{H2}]\text{, }\mathbf{S2\text{:}}[\mathbf{H1}]\Big)+
\text{corr}\Big(\mathbf{S1\text{:}}[\mathbf{B2}]\text{, }\mathbf{S2\text{:}}[\mathbf{H1}]\Big)\bigg).
\end{gathered}
\end{equation*}
Figure \ref{fig:Correlation}.c shows the results of the third experiment, where the proposed method significantly generated better performance in comparison with other HA methods. Indeed, this is the effect of using supervision information in order to provide better functional alignment in MVP analysis. In other words, the proposed method not only maximizes the correlations between the homogeneous (same category) stimuli that are located at the same time points but also it maximizes correlations among the whole of within-class stimulus. In these three experiments, the correlations of neural activities are naturally positive because each time point is only compared to the time points that belong to the same category (class) \cite{tony16,tony17}.

\subsubsection{Correlation analysis across different categories}

The average of correlations for each pair of subjects is generated by using the stimuli, which belong to the different categories. Average of correlations in this experiment is calculated as follows:
\begin{equation}
\begin{split}
\rho_4 = \frac{1}{\Psi_4} \sum_{\substack{i=1\\j = i+1}}^{S}\sum_{\substack{m,n = 1\\n\neq m}}^{L}\sum_{\ell=1}^{L_m}\sum_{k=1}^{L_n}
\text{corr}\Big(\mathbf{X}^{(i)}_{[m;\ell]}\mathbf{R}^{(i)}_{[m;\ell]}\text{, }\mathbf{X}^{(j)}_{[n;k]}\mathbf{R}^{(j)}_{[n;k]}\Big),
\end{split}
\end{equation}
where $\Psi_4=\frac{1}{2}S(S-1)\Big(\sum_{m=1}^{L}\big(L_m\sum_{\substack{n=1\\n\neq m}}^{L}L_n\big)\Big)$. For the example of Figure \ref{fig:DifferentHAs}, we have following comparisons:
\begin{equation*}
\begin{gathered}
\rho_4 = \frac{1}{8}\times\\
\bigg(\text{corr}\Big(\mathbf{S1\text{:}}[\mathbf{H1}]\text{, }\mathbf{S2\text{:}}[\mathbf{B1}]\Big)+
\text{corr}\Big(\mathbf{S1\text{:}}[\mathbf{H1}]\text{, }\mathbf{S2\text{:}}[\mathbf{B2}]\Big)+\\
\text{corr}\Big(\mathbf{S1\text{:}}[\mathbf{H2}]\text{, }\mathbf{S2\text{:}}[\mathbf{B1}]\Big)+
\text{corr}\Big(\mathbf{S1\text{:}}[\mathbf{H2}]\text{, }\mathbf{S2\text{:}}[\mathbf{B2}]\Big)+\\
\text{corr}\Big(\mathbf{S1\text{:}}[\mathbf{B1}]\text{, }\mathbf{S2\text{:}}[\mathbf{H1}]\Big)+
\text{corr}\Big(\mathbf{S1\text{:}}[\mathbf{B1}]\text{, }\mathbf{S2\text{:}}[\mathbf{H2}]\Big)+\\
\text{corr}\Big(\mathbf{S1\text{:}}[\mathbf{B2}]\text{, }\mathbf{S2\text{:}}[\mathbf{H1}]\Big)+
\text{corr}\Big(\mathbf{S1\text{:}}[\mathbf{B2}]\text{, }\mathbf{S2\text{:}}[\mathbf{H2}]\Big)\bigg).
\end{gathered}
\end{equation*}
Figure \ref{fig:Correlation}.d demonstrates the results of the last experiment, where the proposed method and LDHA not only minimize the correlations between different categories of stimuli but also they provide negative correlations in order to discriminate distinctive classes. Since other techniques do not use the supervision information, they cannot set between-class minimization term in the optimization procedure for improving the performance of functional alignment. 

Further, if we compare `NONE' in Figure \ref{fig:Correlation}.b and Figure \ref{fig:Correlation}.d, the correlation is higher for stimuli in different categories than stimuli in the same category before functional alignment. As mentioned before, the neural activities in fMRI datasets are noisy rotations of a shared template across all subjects. In the example of Figure \ref{fig:SHA}, there is a higher between-class correlation among the bottles stimuli in the \emph{Subject 1} and the houses stimuli in the \emph{Subject S} because of those noisy rotations that can increase the undesirable correlation for stimuli in different categories. Indeed, the primary duty of SHA is finding orthogonal vector space for each category of stimuli, where the mapped features in each category have maximum within-class correlations and minimum between-class correlations.

\begin{figure*}[!t] 
	\centering
	\begin{minipage}[b]{0.24\linewidth}
		\includegraphics[width=0.99\textwidth]{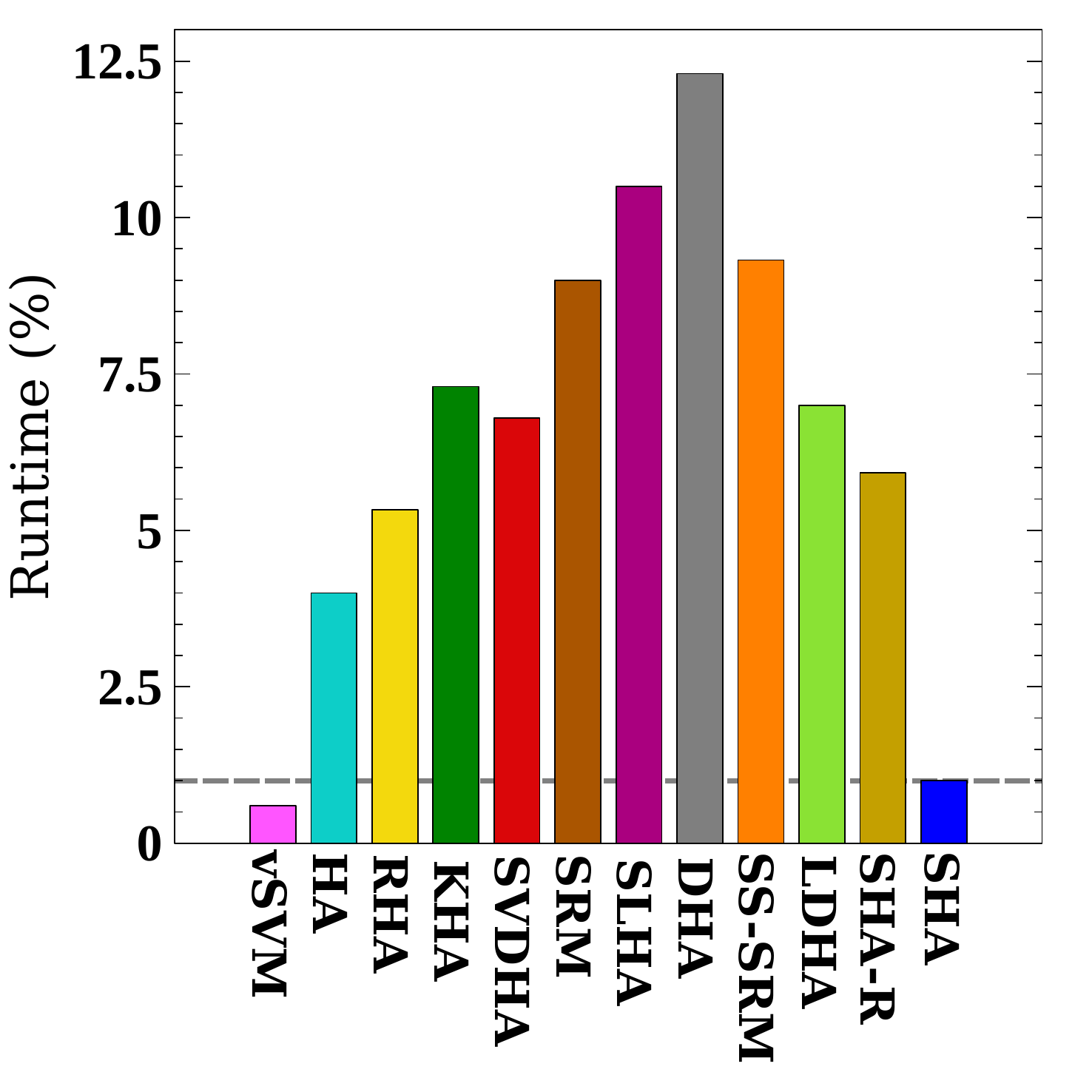}
		\\	\centering {\small (a) DS105}
	\end{minipage}
	\begin{minipage}[b]{0.24\linewidth}
		\includegraphics[width=0.99\textwidth]{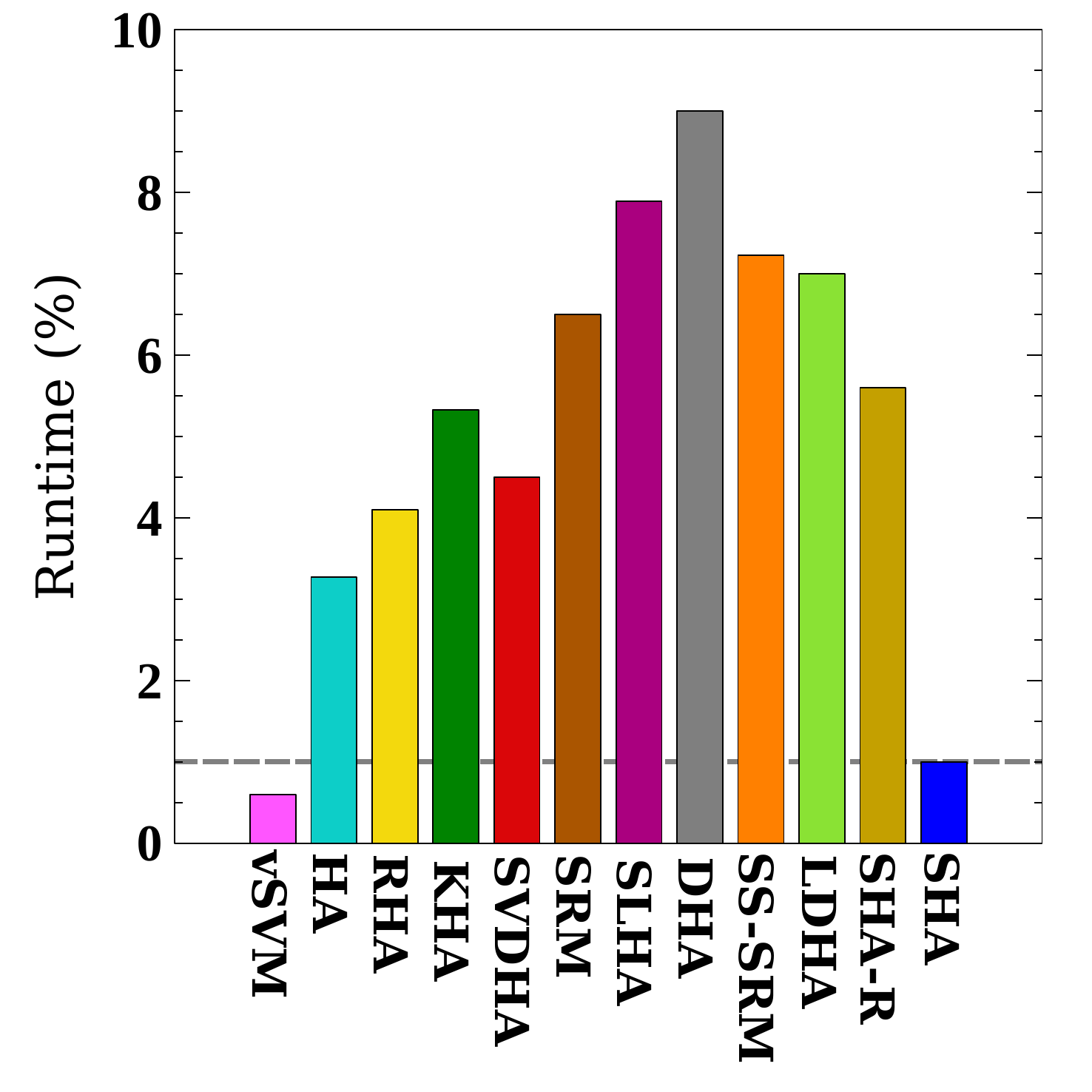}
		\\	\centering {\small (b) DS107}
	\end{minipage}
	\begin{minipage}[b]{0.24\linewidth}
		\includegraphics[width=0.99\textwidth]{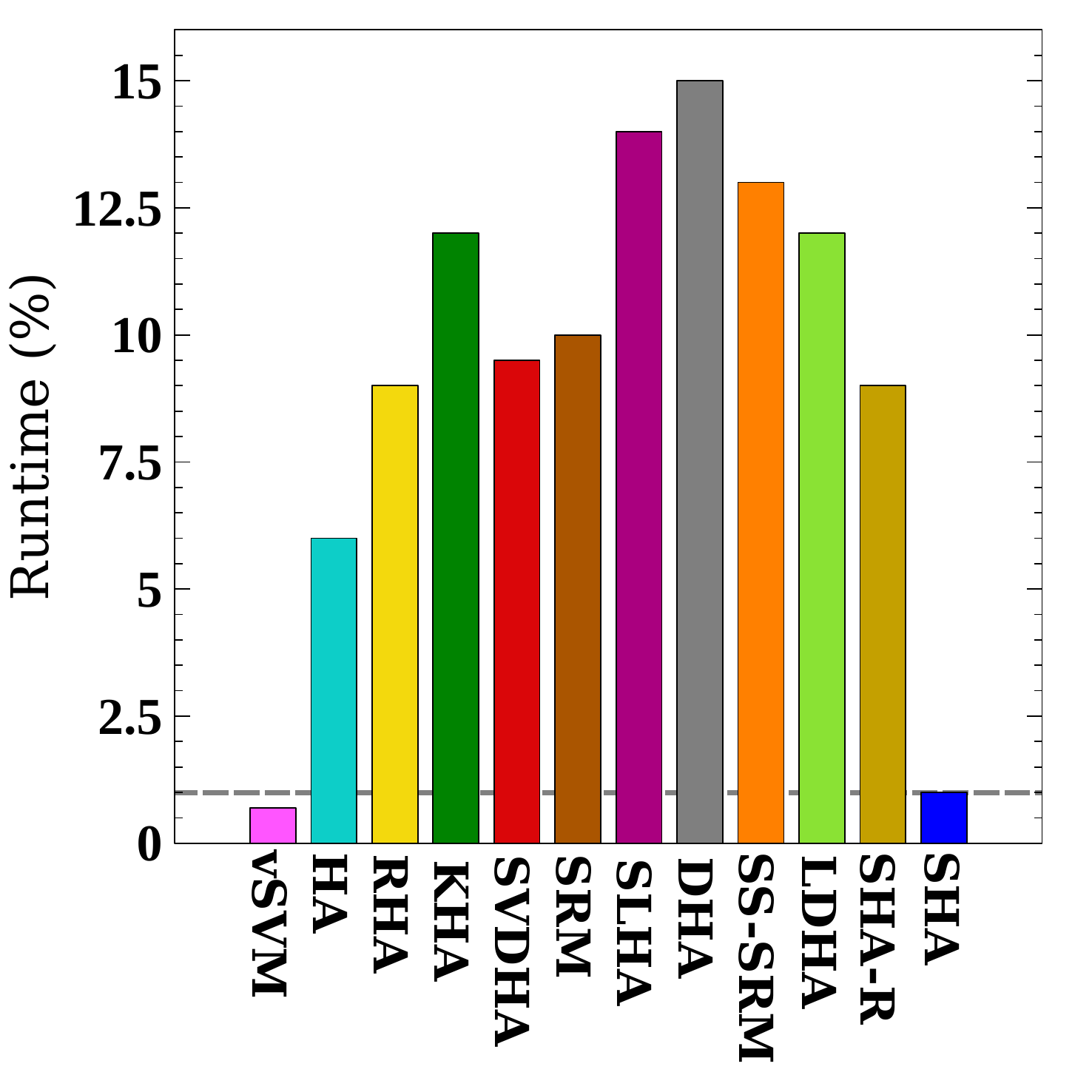}
		\\	\centering {\small (c) CMU}
	\end{minipage}
	\begin{minipage}[b]{0.24\linewidth}
		\includegraphics[width=0.99\textwidth]{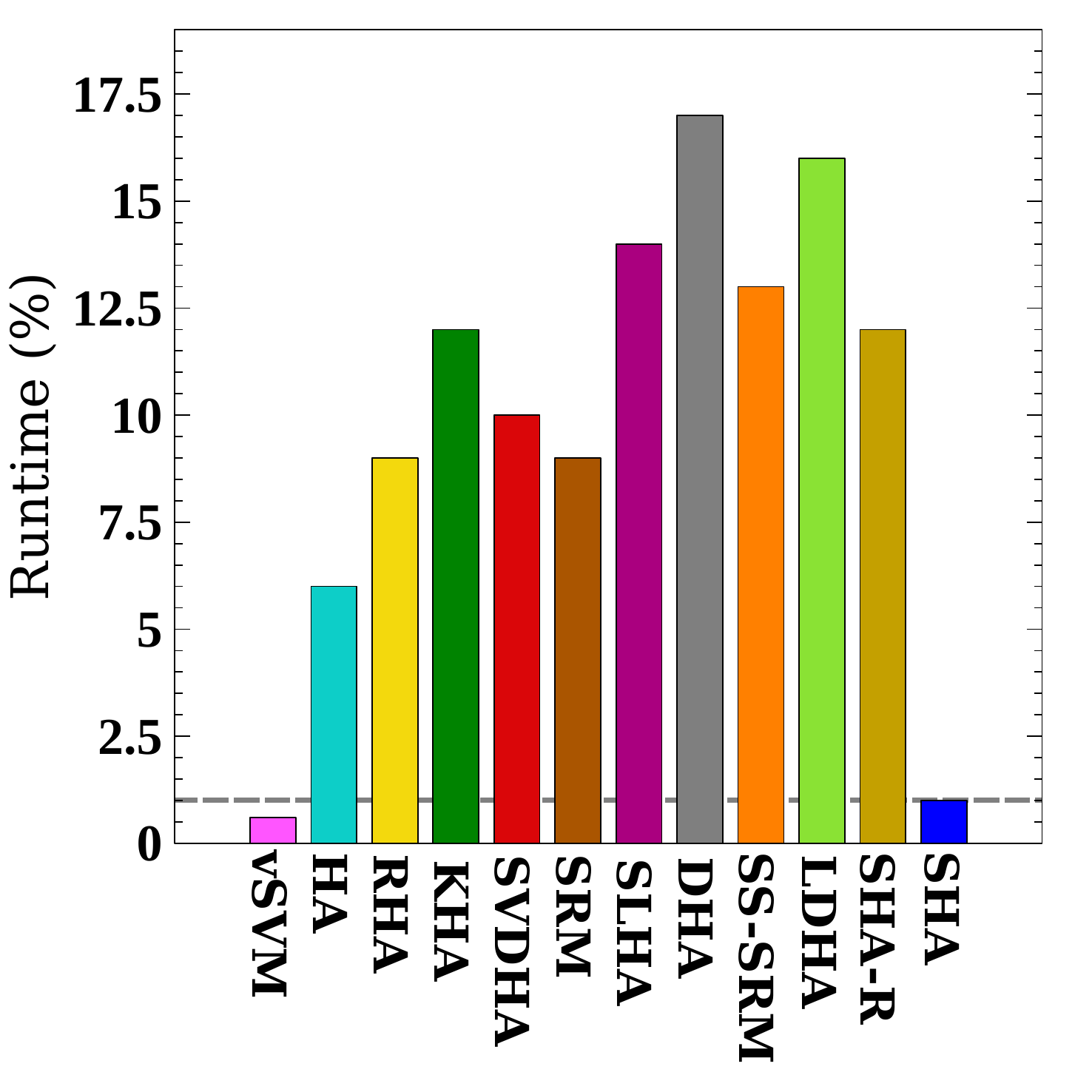}
		\\	\centering {\small (d) Raiders}
	\end{minipage}
	\caption{Comparing runtime of different functional alignment techniques, where the runtime of SHA is considered as a unit}
	\label{fig:Runtime}
	\vskip -0.2in        
\end{figure*}  

\subsection{Performance on Simple Cognitive Tasks}
In the previous section, we show that the proposed method in the training phase can maximize the within-class correlations and minimize the between-class correlations. This section analyzes the performance of different HA methods in the post-alignment classification by using simple tasks datasets that are listed in Table \ref{tbl:Datasets}, i.e., DS005, DS105, DS107, DS116, DS117, and CMU. In fact, these datasets include simple task, e.g., watching a photo \cite{tony16,tony17}. Like previous studies \cite{haxby11,haxby14,xu12,lorbert12,chen14,chen15,chen16,tony17a}, this paper  employs $\nu$-SVM algorithm \cite{smola04} in order to generate the MVP (classification) models in the empirical studies. The binary $\nu$-SVM \cite{smola04} is utilized for datasets with just two classes, i.e., DS005, DS116, and DS117. Further, we use multi-label $\nu$-SVM \cite{smola04,lorbert12} for multi-class datasets, i.e., DS105, DS107, and CMU. In this paper, $\nu$-SVM is employed with fixed parameters (i.e., $\nu=0.5$ and a linear kernel) in order to avoid the effect of training parameters on the performances of evaluated alignment techniques. This paper uses leave-one-subject-out cross-validation for evaluating the generated models, where the neural activities belong to all subjects except one is iteratively selected as the training-set, and the neural responses of the unselected subject are considered as the testing-set for that iteration. As the first step, training-set are applied to different HA methods. Then, the aligned training-set is used for generating the classification models. Finally, the testing-set are employed for evaluating both the functional alignment and classification analysis. It is worth noting that the same preprocessed datasets are applied to all hyperalignment methods in each iteration of the cross validation. In other words, we consider whole of the classification procedure is fixed except the alignment section, i.e., the same preprocessed dataset for each iteration, the same learning algorithm, the same parameters for learning, etc.

Figure \ref{fig:ClassificationAccuracy}.a illustrates the accuracy of binary classification analysis. In addition, Figure \ref{fig:ClassificationAccuracy}.b shows the classification accuracy for multi-class datasets. As depicted in these figures, the accuracy of the preprocessed datasets without functional alignment is limited in comparison with the aligned data. Since SHA employs the supervision information in order to align the neural activities, it achieves superior performance in comparison with other methods. The effect of supervision information on the performance of functional alignment is more highlight on multi-class datasets.

\subsection{Performance on Complex Cognitive Tasks}
This section compares the performance of HA methods by employing two fMRI datasets related to movie stimuli, i.e., DS113 and Raiders. The scheme of experiments in this section is exactly same as the previous section, i.e., the cross-validation, the multi-class $\nu$-SVM, etc. Figure \ref{fig:Movies} depicts the generated results, where `$\nu$-SVM' is the classification analysis without functional alignment. In this section, we have utilized the method proposed in \cite{haxby11} for ranking the voxels in ROI by using their neurological priorities \cite{xu12,chen14,chen15}. Indeed, different number of voxels in each hemisphere are selected as follows for generating the experiments: $[100, 200, 400, 600, 800, 1000, 1200]$ for DS113 dataset, and $[70, 140, 210, 280, 350, 420, 490]$ for Raiders dataset. Further, these experiments are repeated by employing the different numbers of time points, i.e., the first $100,  400, 800,\text{ and }2000$ TRs in both datasets. Figure \ref{fig:Movies} demonstrates that SHA has generated superior performance in comparison with other HA algorithms. The effect of supervision information on the performance of the proposed method can be more significant when the number of time points is limited.

\subsection{Time Complexity and Runtime Analysis}
As mentioned before, SHA can improve runtime of alignment procedure by applying a single iteration optimization approach. SHA has two main steps, i.e., applying SVD decomposition to the subject neural activities, and calculating the supervised shared space by using Incremental PCA. The rest of algorithm is simple matrix operations such as sum and product. The time complexity for the first SVD decomposition in SHA is $\mathcal{O}\big((T+L)VS\big)$, where $\mathcal{O}()$ is the `big O' time complexity, $S$ denotes the number of subjects, $T$ is the time points, $L$ denotes the number of categories, and $V$ is the number of voxels. Like most of other HA approaches, the first SVD decomposition on subjects' data can be parallelized; thus it can reduce the time complexity to $\mathcal{O}\big(\zeta^{-1}(T+L)VS\big)$, where $\zeta$ is the number of parallel cores. Further, applying Incremental PCA as the second step has only $\mathcal{O}\big(L^2S\big)$ time complexity. It is worth noting that in most of fMRI problems, we have: $L, S \leq T \leq V$. These time complexity can significantly increase the runtime of SHA in comparison with other state-of-the-art techniques, e.g., the time complexity of SRM in BrainIAK library is $\mathcal{O}\big(\zeta^{-1}\tau S T V\big)$, where $\tau$ is the number of iterations \cite{turek16,Anderson16}.

Runtime of the proposed SHA has been compared with other HA methods, by utilizing following datasets: DS105, DS107, CMU, and Raiders. As previously mentioned, a PC with certain specifications ($8$ cores and $16$ threads) is employed in order to generate all of the empirical studies. Figure \ref{fig:Runtime} compares the runtime of SHA with other functional alignment methods, where all runtime are scaled based on the proposed method, i.e., the runtime of SHA is used as a unit. All of the algorithms in this paper employ one of the parallel processes libraries such as Message Passing Interface (MPI) (that is used by SRM, SS-SRM, etc.) in order to align different subjects neural activities. SHA and DHA are implemented by Pytorch; however, we reported the runtime, where Pytorch library only uses CPU (no access to GPU). As these figures illustrate, the runtime of DHA is the worse one in comparison with other techniques because training deep networks increase the time complexity of alignment. After that, SHLA also cannot provide acceptable runtime because of the ensemble approach. Since LDHA, SS-SRM, and SRM must iteratively align the neural activities in order to seek the optimum solution, their time complexities are high. Based on the performance of SHA in the previous sections, the proposed method produces acceptable runtime because the optimization procedure in SHA does not need any iteration. 

\begin{figure*}[!t]
	\begin{center}
		\includegraphics[width=0.8\textwidth]{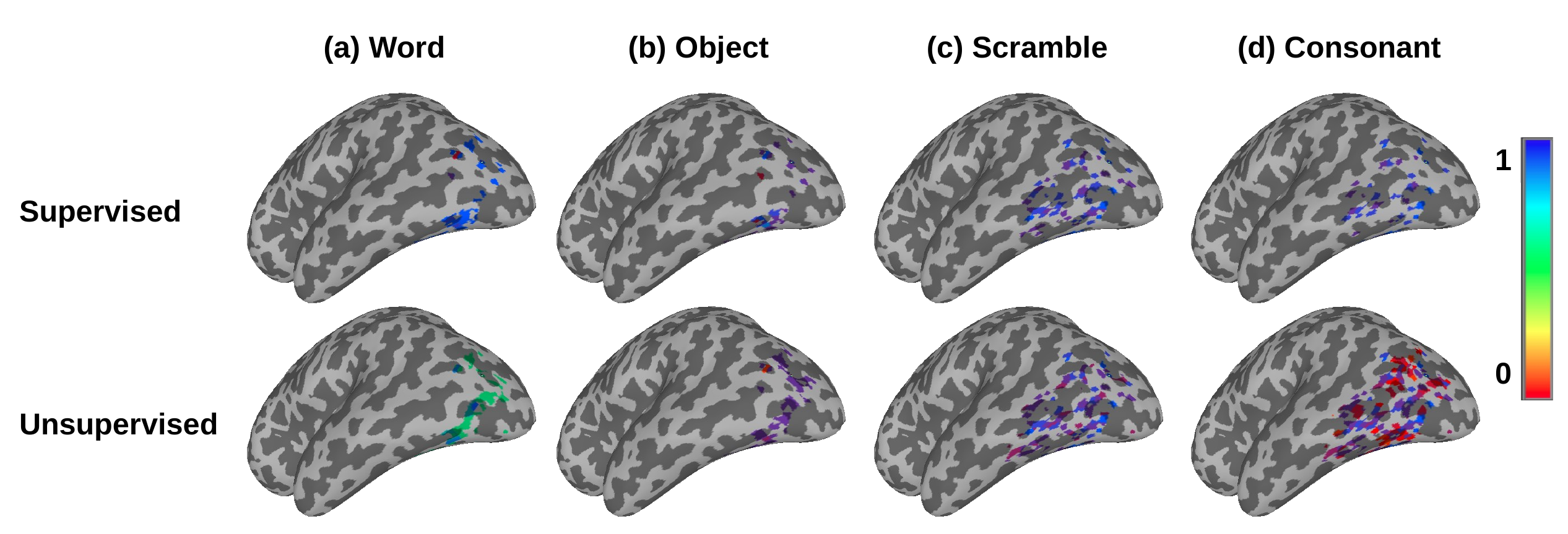}\\ 
		\caption{Comparing the neural activities in dataset DS107: supervised shared space ($\mathbf{W}$) vs. unsupervised shared space ($\mathbf{G}$)}
		\label{fig:VisualizingW}
	\end{center}
	\vskip -0.3in
\end{figure*}

\subsection{Comparing shared spaces}
This section uses dataset DS107 to generate and then compare both the supervised shared space ($\mathbf{W}$) and the unsupervised shared space ($\mathbf{G}$). Figure~\ref{fig:VisualizingW} illustrates the generate neural activities by using SHA. Here, the unsupervised shared space is visualized by averaging the neural activities in each category of stimuli. In this dataset, we visualized the neural activities located on the left side of the brain because DS107 has no ROI on the right side \cite{duncan09}. As this figure showed, the neural activities are significantly distinctive across categories of stimuli. For instance, word stimuli and object category have a strong correlation as the meaningful concepts, whereas the neural signatures of these categories are completely different from non-meaningful concepts, i.e., consonant and the scramble stimuli. Similarly, the non-meaningful concepts also have a strong correlation in dataset DS107. Moreover, visualizing the supervised shared space can show that SHA has selected exclusively unique shared information for each category of stimuli by leveraging the supervision information in the alignment procedure. While the supervised shared space only includes the activation with a higher probability of happening in all time points belonging to a unique category of stimuli, the average of neural activities in the unsupervised shared space can represent more general signatures.

\section{Conclusion}
This paper has developed Supervised Hyperalignment (SHA) for improving the performance of the hyperalignment methods in supervised fMRI analysis, i.e., MVP problems. Further, we have also demonstrated how SHA can be utilized for post-alignment classification with both simple tasks and complex tasks datasets. SHA can maximize within-class correlation and minimize the between-class correlation, whereas unsupervised methods only maximize the correlation between the voxels with the same position in the time series. SHA can also handle large datasets because it does not reference the training-set in testing-stage and uses a generalized solution (including rank-$\mu$ SVD and Incremental PCA) for optimization. Empirical studies confirm that the runtime of SHA is almost the same as only a single iteration of prevalent unsupervised alignment techniques such as Shared Response Model (SRM) or Regularized/Kernel Hyperalignment. Further, experiments on multi-subject datasets demonstrate that SHA method achieves superior performance (i.e., up to 19\% for multi-class datasets) in comparison with other state-of-the-art HA algorithms. In the future, we will plan to develop a deep version of SHA to improve its performance in nonlinear problems.

\section*{Proofs}
\textbf{Lemma 1}\\
\emph{Proof.} By considering the constraint, Equation (8) can be written as follows:
\begin{equation*}
\begin{gathered}
\underset{\mathbf{R}^{(i)}, \mathbf{R}^{(j)}}{\min}\Bigg\{\sum_{i = 1}^{S}\sum_{j = i+1}^{S}\| \mathbf{K}^{(i)}\mathbf{X}^{(i)}\mathbf{R}^{(i)} - \mathbf{K}^{(j)}\mathbf{X}^{(j)}\mathbf{R}^{(j)} \|^2_F\Bigg\} =\\
\underset{\mathbf{R}^{(i)}, \mathbf{R}^{(j)}}{\min}\Bigg\{\sum_{i = 1}^{S}\sum_{j = i+1}^{S}\bigg(
\|\mathbf{K}^{(i)}\mathbf{X}^{(i)}\mathbf{R}^{(i)}\|^2_F + \|\mathbf{K}^{(j)}\mathbf{X}^{(j)}\mathbf{R}^{(j)}\|^2_F\bigg)\\
- 2 \sum_{i = 1}^{S}\sum_{j = i+1}^{S} \text{ tr}\bigg(\big(\mathbf{K}^{(i)}\mathbf{X}^{(i)}\mathbf{R}^{(i)}\big)^\top \mathbf{K}^{(j)}\mathbf{X}^{(j)}\mathbf{R}^{(j)}\bigg)\Bigg\}=\\
\underset{\mathbf{R}^{(i)}, \mathbf{R}^{(j)}}{\min}\Bigg\{\sum_{i = 1}^{S}\sum_{j = i+1}^{S}\bigg( 2 \text{ tr}\big(\mathbf{I}_{V}\big) -
\epsilon\|\mathbf{R}^{(i)}\|^2_F - \epsilon\|\mathbf{R}^{(j)}\|^2_F \bigg)\\
- 2 \sum_{i = 1}^{S}\sum_{j = i+1}^{S} \text{ tr}\bigg(\big(\mathbf{K}^{(i)}\mathbf{X}^{(i)}\mathbf{R}^{(i)}\big)^\top \mathbf{K}^{(j)}\mathbf{X}^{(j)}\mathbf{R}^{(j)}\bigg)\Bigg\}=\\
\underset{\mathbf{R}^{(i)}, \mathbf{R}^{(j)}}{\min}\Bigg\{ VS (S - 1) - \epsilon \sum_{i = 1}^{S}\sum_{j = i+1}^{S}\bigg(\|\mathbf{R}^{(i)}\|^2_F + \|\mathbf{R}^{(j)}\|^2_F\bigg)\\
- 2 \sum_{i = 1}^{S}\sum_{j = i+1}^{S} \text{ tr}\bigg(\big(\mathbf{K}^{(i)}\mathbf{X}^{(i)}\mathbf{R}^{(i)}\big)^\top \mathbf{K}^{(j)}\mathbf{X}^{(j)}\mathbf{R}^{(j)}\bigg)\Bigg\}=\\
\underset{\mathbf{R}^{(i)}, \mathbf{R}^{(j)}}{\min}\Bigg\{ VS (S - 1) \\- \epsilon \sum_{i = 1}^{S}\bigg((S - i)\|\mathbf{R}^{(i)}\|^2_F + (i - 1)\|\mathbf{R}^{(i)}\|^2_F\bigg)\\
- 2 \sum_{i = 1}^{S}\sum_{j = i+1}^{S} \text{ tr}\bigg(\big(\mathbf{K}^{(i)}\mathbf{X}^{(i)}\mathbf{R}^{(i)}\big)^\top \mathbf{K}^{(j)}\mathbf{X}^{(j)}\mathbf{R}^{(j)}\bigg)\Bigg\}=\\
\underset{\mathbf{R}^{(i)}, \mathbf{R}^{(j)}}{\min}\Bigg\{ VS (S - 1) - \epsilon (S - 1) \sum_{i = 1}^{S}\|\mathbf{R}^{(i)}\|^2_F \\
\end{gathered}
\end{equation*}
\begin{equation*}
\begin{gathered}
- 2 \sum_{i = 1}^{S}\sum_{j = i+1}^{S} \text{ tr}\bigg(\big(\mathbf{K}^{(i)}\mathbf{X}^{(i)}\mathbf{R}^{(i)}\big)^\top \mathbf{K}^{(j)}\mathbf{X}^{(j)}\mathbf{R}^{(j)}\bigg)\Bigg\} \equiv\\
\underset{\mathbf{R}^{(i)}, \mathbf{R}^{(j)}}{\max}\Bigg\{\frac{2}{S - 1}\sum_{i = 1}^{S}\sum_{j = i+1}^{S}\text{tr}\Big((\mathbf{K}^{(i)}\mathbf{X}^{(i)}\mathbf{R}^{(i)})^\top\mathbf{K}^{(j)}\mathbf{X}^{(j)}\mathbf{R}^{(j)}\Big)\\
+ \epsilon \sum_{\ell=1}^{S} \bigg\| \mathbf{R}^{(\ell)}\bigg\|_F^2\Bigg\} \quad\blacksquare
\end{gathered}
\end{equation*}
\textbf{Lemma 2}\\
\emph{Proof.} Equation (7) can be written as the trace form:
\begin{equation*}
\begin{gathered}
\underset{\mathbf{R}^{(i)}, \mathbf{R}^{(j)}}{\min}\Bigg\{\sum_{i = 1}^{S}\sum_{j = i+1}^{S}\bigg\| \mathbf{K}^{(i)}\mathbf{X}^{(i)}\mathbf{R}^{(i)} - \mathbf{K}^{(j)}\mathbf{X}^{(j)}\mathbf{R}^{(j)} \bigg\|^2_F\Bigg\} =
\end{gathered}
\end{equation*}
\begin{equation*}
\begin{gathered}
\underset{\mathbf{R}^{(i)}, \mathbf{R}^{(j)}}{\min}\Bigg\{\frac{1}{2}\sum_{i = 1}^{S}\sum_{j = 1}^{S}\bigg\| \mathbf{K}^{(i)}\mathbf{X}^{(i)}\mathbf{R}^{(i)} - \mathbf{K}^{(j)}\mathbf{X}^{(j)}\mathbf{R}^{(j)} \bigg\|^2_F\Bigg\} =\\
\underset{\mathbf{R}^{(i)}, \mathbf{R}^{(j)}, \mathbf{W}}{\min}\Bigg\{\frac{1}{2}\sum_{i = 1}^{S}\sum_{j = 1}^{S}\bigg\| \Big( \mathbf{K}^{(i)}\mathbf{X}^{(i)}\mathbf{R}^{(i)} - \mathbf{W} \Big) + \\\Big( \mathbf{W} - \mathbf{K}^{(j)}\mathbf{X}^{(j)}\mathbf{R}^{(j)} \Big) \bigg\|^2_F\Bigg\} =\\
\underset{\mathbf{R}^{(i)}, \mathbf{R}^{(j)}, \mathbf{W}}{\min}\Bigg\{\frac{1}{2}\sum_{i = 1}^{S}\sum_{j = 1}^{S}\bigg\| \mathbf{K}^{(i)}\mathbf{X}^{(i)}\mathbf{R}^{(i)} - \mathbf{W} \bigg\|^2_F + \\\frac{1}{2}\sum_{i = 1}^{S}\sum_{j = 1}^{S}\bigg\| \mathbf{W} - \mathbf{K}^{(j)}\mathbf{X}^{(j)}\mathbf{R}^{(j)}\bigg\|^2_F - \\
 \sum_{i = 1}^{S}\sum_{j = 1}^{S}\text{ tr}\bigg( \Big( \mathbf{K}^{(i)}\mathbf{X}^{(i)}\mathbf{R}^{(i)} - \mathbf{W} \Big)^\top\\
\Big( \mathbf{W} - \mathbf{K}^{(j)}\mathbf{X}^{(j)}\mathbf{R}^{(j)} \Big) \bigg)\Bigg\}=\\
\underset{\mathbf{R}^{(i)}, \mathbf{R}^{(j)}, \mathbf{W}}{\min}\Bigg\{S\sum_{i = 1}^{S}\bigg\| \mathbf{K}^{(i)}\mathbf{X}^{(i)}\mathbf{R}^{(i)} - \mathbf{W} \bigg\|^2_F - \\
\text{ tr}\Bigg( \bigg(\sum_{i = 1}^{S}\Big(\mathbf{K}^{(i)}\mathbf{X}^{(i)}\mathbf{R}^{(i)}\Big) - S\mathbf{W} \bigg)^\top\\
\bigg( S\mathbf{W} - \sum_{j = 1}^{S}\Big(\mathbf{K}^{(j)}\mathbf{X}^{(j)}\mathbf{R}^{(j)}\Big)\bigg) \Bigg)\Bigg\}\text{, }\\
\end{gathered}
\end{equation*}
Here, based on the definition of $\mathbf{W}$, both terms $\bigg(\sum_{i = 1}^{S}\Big( \mathbf{K}^{(i)}\mathbf{X}^{(i)}\mathbf{R}^{(i)}\Big) - S\mathbf{W} \bigg)^\top = 0$ and $S\mathbf{W} - \sum_{j = 1}^{S}\Big(\mathbf{K}^{(j)}\mathbf{X}^{(j)}\mathbf{R}^{(j)}\Big) = 0$; thus we have:
\begin{equation*}
\begin{gathered}
\underset{\mathbf{R}^{(i)}, \mathbf{W}}{\min}\Bigg\{S\sum_{i = 1}^{S}\bigg\| \mathbf{K}^{(i)}\mathbf{X}^{(i)}\mathbf{R}^{(i)} - \mathbf{W} \bigg\|^2_F \Bigg\}\\
\equiv\underset{\mathbf{R}^{(i)}, \mathbf{W}}{\min}\Bigg\{\sum_{i = 1}^{S} \bigg\| \mathbf{K}^{(i)}\mathbf{X}^{(i)}\mathbf{R}^{(i)} -  \mathbf{W} \bigg\|^2_F\Bigg\} \quad\blacksquare
\end{gathered}
\end{equation*}
\textbf{Lemma 3}\\
\emph{Proof.} 
Since equality possesses the transitive property, it is enough that we prove:
\begin{equation*}
\begin{gathered}
\min_{x \in \mathbb{R}^N} \min_{y \in \mathbb{R}^M}  f(x,y)  = \min_{(x,y) \in \mathbb{R}^N \times \mathbb{R}^M}  f(x,y)
\end{gathered}
\end{equation*}
and
\begin{equation*}
\begin{gathered}
\min_{y \in \mathbb{R}^M} \min_{x \in \mathbb{R}^N}  f(x,y) = \min_{(x,y) \in \mathbb{R}^N \times \mathbb{R}^M}  f(x,y),
\end{gathered}
\end{equation*}
and by symmetry of the notation, it is enough that we prove
\begin{equation*}
\begin{gathered}
\min_{x \in \mathbb{R}^N} \min_{y \in \mathbb{R}^M}  f(x,y)  = \min_{(x,y) \in \mathbb{R}^N \times \mathbb{R}^M}  f(x,y).
\end{gathered}
\end{equation*}
We will prove this equality in two steps. 
\\ \vspace{1mm} \\
\underline{Step 1}:
\begin{equation*}
\begin{gathered}
\min_{x \in \mathbb{R}^N} \min_{y \in \mathbb{R}^M}  f(x,y)   \geq \min_{(x,y) \in \mathbb{R}^N \times \mathbb{R}^M}  f(x,y).
\end{gathered}
\end{equation*}
Since $\mathbb{R}^M$ is isomorphic, homeomorphic and diffeomeorphic to the subspaces $\{ c \} \times \mathbb{R}^M$ for every $c \in \mathbb{R}^N$, then
\begin{equation*}
\begin{gathered}
\min_{y \in \mathbb{R}^M}  f(x,y)   \geq \min_{(x,y) \in \mathbb{R}^N \times \mathbb{R}^M}  f(x,y).
\end{gathered}
\end{equation*}
Morally $ \mathbb{R}^M  \subset \mathbb{R}^N \times \mathbb{R}^M$ and so since we are taking the minimum over a smaller space, the minimum is bigger, by the very definition of minimum.  Note that the right hand side of this inequality is independent of $x$, and so we can take to both sides of the last inequality $\min_{x \in \mathbb{R}^N}$ to obtain:
\begin{equation*}
\begin{gathered}
\min_{x \in \mathbb{R}^N} \min_{y \in \mathbb{R}^M}  f(x,y)   \geq \min_{x \in \mathbb{R}^N} \min_{(x,y) \in \mathbb{R}^N \times \mathbb{R}^M}  f(x,y)=\\
\min_{(x,y) \in \mathbb{R}^N \times \mathbb{R}^M}  f(x,y).
\end{gathered}
\end{equation*}
This proves Step 1.
\\ \vspace{1mm} \\
\underline{Step 2}:
\begin{equation*}
\begin{gathered}
\min_{x \in \mathbb{R}^N} \min_{y \in \mathbb{R}^M}  f(x,y)   \leq \min_{(x,y) \in \mathbb{R}^N \times \mathbb{R}^M}  f(x,y).
\end{gathered}
\end{equation*}
Suppose that the statement in  Step 2 is not true, namely that the inequality in Step 1 is strict, then
\begin{equation*}
\begin{gathered}
\min_{x \in \mathbb{R}^N} \min_{y \in \mathbb{R}^M}  f(x,y) > \min_{(x,y) \in \mathbb{R}^N \times \mathbb{R}^M}  f(x,y).
\end{gathered}
\end{equation*}
This is an inequality between two real numbers:
\begin{equation*}
\begin{gathered}
M_1:=\min_{x \in \mathbb{R}^N} \min_{y \in \mathbb{R}^M}  f(x,y)
\end{gathered}
\end{equation*}
and
\begin{equation*}
\begin{gathered}
M_2:=\min_{(x,y) \in \mathbb{R}^N \times \mathbb{R}^M}  f(x,y)
\end{gathered}
\end{equation*}
and therefore, since they are different by hypothesis of absurd, they are at a positive Euclidean distance:
\begin{equation*}
\begin{gathered}
\delta:=M_1-M_2>0.
\end{gathered}
\end{equation*}
Therefore, we have:
\begin{equation*}
\begin{gathered}
\min_{x \in \mathbb{R}^N} \min_{y \in \mathbb{R}^M}  f(x,y) > \frac{\delta}{2}+ \min_{(x,y) \in \mathbb{R}^N \times \mathbb{R}^M}  f(x,y).
\end{gathered}
\end{equation*}
By definition of $\min$, we have:
\begin{equation*}
\begin{gathered}
f(x,y) \geq \min_{y \in \mathbb{R}^M}  f(x,y) \geq \min_{x \in \mathbb{R}^N} \min_{y \in \mathbb{R}^M}  f(x,y)
\end{gathered}
\end{equation*}
and so
\begin{equation*}
\begin{gathered}
f(x,y)  > \frac{\delta}{2}+ \min_{(x,y) \in \mathbb{R}^N \times \mathbb{R}^M}  f(x,y).
\end{gathered}
\end{equation*}
Note that the right hand side of this inequality is independent of $x$ and $y$, and so I can take to both sides $\min_{(x,y) \in \mathbb{R}^N \times \mathbb{R}^M}$ to obtain:
\begin{equation*}
\begin{gathered}
\min_{(x,y) \in \mathbb{R}^N \times \mathbb{R}^M} f(x,y)  >\\
\min_{(x,y) \in \mathbb{R}^N \times \mathbb{R}^M} \left \{ \frac{\delta}{2}+ \min_{(x,y) \in \mathbb{R}^N \times \mathbb{R}^M}  f(x,y) \right \}=\\
\frac{\delta}{2}+ \min_{(x,y) \in \mathbb{R}^N \times \mathbb{R}^M} \left \{  \min_{(x,y) \in \mathbb{R}^N \times \mathbb{R}^M}  f(x,y) \right \}=\\
\frac{\delta}{2}+ \min_{(x,y) \in \mathbb{R}^N \times \mathbb{R}^M}  f(x,y).
\end{gathered}
\end{equation*}
Using our notation, we have
\begin{equation*}
\begin{gathered}
M_2>\frac{\delta}{2}+M_2,
\end{gathered}
\end{equation*}
and so 
\begin{equation*}
\begin{gathered}
0> \frac{\delta}{2},
\end{gathered}
\end{equation*}
which is a contradiction because by hypothesis of absurd, $\delta>0$. This concludes the proof of Step 2 and so the proof of the Lemma 3, since $M_1 \geq M_2$ and $M_2 \geq M_1$ imply $M_1=M_2$.$\quad\blacksquare$\\
\textbf{Lemma 4}\\
\emph{Proof.} 
\begin{equation*}
\begin{split}
\underset{\mathbf{W}, \mathbf{R}^{(i)} }{\min}\bigg\{\sum_{i = 1}^{S} \| \mathbf{W} - \mathbf{K}^{(i)}\mathbf{X}^{(i)}\mathbf{R}^{(i)}  \|^2_F\bigg\}=
\end{split}
\end{equation*}
By substituting $\mathbf{R}^{(\ell)}=
\Big(\big(\mathbf{K}^{(\ell)}\mathbf{X}^{(\ell)}\big)^\top\mathbf{K}^{(\ell)}\mathbf{X}^{(\ell)} + \epsilon\mathbf{I}_V\Big)^{-1}\big(\mathbf{K}^{(\ell)}\mathbf{X}^{(\ell)}\big)^\top\mathbf{W}$, we have: 
\begin{equation*}
\begin{gathered}
\underset{\mathbf{W}, \mathbf{R}^{(i)} }{\min}\bigg\{\sum_{i = 1}^{S}\Big\| \mathbf{W}  -\\\mathbf{K}^{(i)}\mathbf{X}^{(i)}\Big(\big(\mathbf{K}^{(i)}\mathbf{X}^{(i)}\big)^\top\mathbf{K}^{(i)}\mathbf{X}^{(i)} + \epsilon\mathbf{I}_V\Big)^{-1}\big(\mathbf{K}^{(i)}\mathbf{X}^{(i)}\big)^\top\mathbf{W}  \Big\|^2_F\bigg\}
\end{gathered}
\end{equation*}
Based on equation (14), we have:
\begin{equation*}
\begin{gathered}
\underset{\mathbf{W}}{\min}\bigg\{\sum_{i = 1}^{S} \Big\| \mathbf{W} - \mathbf{P}^{(i)}\mathbf{W}  \Big\|^2_F\bigg\}\\
=\underset{\mathbf{W}}{\min}\bigg\{\sum_{i = 1}^{S} \Big\| \big(\mathbf{I}_{L} - \mathbf{P}^{(i)}\big)\mathbf{W}  \Big\|^2_F\bigg\}\\
=\underset{\mathbf{W}}{\min}\bigg\{\sum_{i=1}^{S} \text{tr}\bigg(\Big(\big(\mathbf{I}_{L} - \mathbf{P}^{(i)}\big)\mathbf{W}\Big)^\top\big(\mathbf{I}_{L} - \mathbf{P}^{(i)}\big)\mathbf{W} \bigg)\bigg\}\\
=\underset{\mathbf{W}}{\min}\bigg\{\sum_{i=1}^{S} \text{tr}\bigg(\mathbf{W}^\top\big(\mathbf{I}_{L} - \mathbf{P}^{(i)}\big)^\top\big(\mathbf{I}_{L} - \mathbf{P}^{(i)}\big)\mathbf{W} \bigg)\bigg\}\\
=\underset{\mathbf{W}}{\min}\bigg\{\sum_{i=1}^{S} \text{tr}\bigg(\mathbf{W}^\top\big(\mathbf{I}_{L} - \mathbf{P}^{(i)}\big)^2\mathbf{W} \bigg)\bigg\}\\
=\underset{\mathbf{W}}{\min}\bigg\{\sum_{i=1}^{S} \text{tr}\bigg(\mathbf{W}^\top\Big(\mathbf{I}_{L}^2
+ \big(\mathbf{P}^{(i)}\big)^2 - 2\mathbf{I}_{L}\mathbf{P}^{(i)}\Big)\mathbf{W} \bigg)\bigg\}
\end{gathered}
\end{equation*}
Since $\mathbf{P}^{(i)}$ is idempotent ($\big(\mathbf{P}^{(i)}\big)^2=\mathbf{P}^{(i)}$) [25], [26], we have:
\begin{equation*}
\begin{gathered}
=\underset{\mathbf{W}}{\min}\bigg\{\sum_{i=1}^{S} \text{tr}\bigg(\mathbf{W}^\top\Big(\mathbf{I}_{L}^2 + \mathbf{P}^{(i)} - 2\mathbf{P}^{(i)}\Big)\mathbf{W} \bigg)\bigg\}\\
\end{gathered}
\end{equation*}
\begin{equation*}
\begin{gathered}
=\underset{\mathbf{W}}{\min}\bigg\{\sum_{i=1}^{S} \text{tr}\bigg(\mathbf{W}^\top\big(\mathbf{I}_L - \mathbf{P}^{(i)}\big)\mathbf{W} \bigg)\bigg\}\\
=\underset{\mathbf{W}}{\min}\Bigg\{\text{tr}\bigg(\mathbf{W}^\top\big(\sum_{i=1}^{S}\mathbf{I}_L - \mathbf{P}^{(i)}\big)\mathbf{W} \bigg)\Bigg\}\\
\equiv \underset{\mathbf{W}}{\min}\Big\{\text{tr}\big(\mathbf{W}^\top\mathbf{U}\mathbf{W}\big)\Big\}\quad\blacksquare
\end{gathered}
\end{equation*}

\section*{Conflict of Interests }
All authors declare that they have no conflict of interest.
\section*{Ethical Approval}
This article does not contain any studies with human participants or animals performed by any of the authors.

\section*{Acknowledgments}
This work was supported by the National Natural Science Foundation of China (Nos. 61876082, 61732006, 61861130366), the National Key R\&D Program of China (Grant Nos. 2018YFC2001600, 2018YFC2001602, 2018ZX10201002), the
Alberta Machine Intelligence Institute (Amii), and the Royal Society-Academy of Medical Sciences Newton Advanced Fellowship (No. NAF$\backslash$R1$\backslash$180371). 


\bibliographystyle{IEEEtran}
\bibliography{SHA}

\begin{thebibliography}{10}
\providecommand{\url}[1]{#1}
\csname url@samestyle\endcsname
\providecommand{\newblock}{\relax}
\providecommand{\bibinfo}[2]{#2}
\providecommand{\BIBentrySTDinterwordspacing}{\spaceskip=0pt\relax}
\providecommand{\BIBentryALTinterwordstretchfactor}{4}
\providecommand{\BIBentryALTinterwordspacing}{\spaceskip=\fontdimen2\font plus
\BIBentryALTinterwordstretchfactor\fontdimen3\font minus
  \fontdimen4\font\relax}
\providecommand{\BIBforeignlanguage}[2]{{%
\expandafter\ifx\csname l@#1\endcsname\relax
\typeout{** WARNING: IEEEtran.bst: No hyphenation pattern has been}%
\typeout{** loaded for the language `#1'. Using the pattern for}%
\typeout{** the default language instead.}%
\else
\language=\csname l@#1\endcsname
\fi
#2}}
\providecommand{\BIBdecl}{\relax}
\BIBdecl

\bibitem{haxby11}
J.~V. Haxby, J.~S. Guntupalli, A.~C. Connolly, Y.~O. Halchenko, B.~R. Conroy,
  M.~I. Gobbini, M.~Hanke, and P.~J. Ramadge, ``A common, high-dimensional
  model of the representational space in human ventral temporal cortex,''
  \emph{Neuron}, vol.~72, no.~2, pp. 404--416, 2011.

\bibitem{haxby14}
J.~V. Haxby, A.~C. Connolly, and J.~S. Guntupalli, ``Decoding neural
  representational spaces using multivariate pattern analysis,'' \emph{Annual
  Review of Neuroscience}, vol.~37, pp. 435--456, 2014.

\bibitem{diedrichsen17}
J.~Diedrichsen and N.~Kriegeskorte, ``Representational models: A common
  framework for understanding encoding, pattern-component, and
  representational-similarity analysis,'' \emph{PLoS Computational Biology},
  vol.~13, no.~4, p. e1005508, 2017.

\bibitem{xu12}
H.~Xu, A.~Lorbert, P.~J. Ramadge, J.~S. Guntupalli, and J.~V. Haxby,
  ``Regularized hyperalignment of multi-set fmri data,'' in \emph{Statistical
  Signal Processing Workshop (SSP)}.\hskip 1em plus 0.5em minus 0.4em\relax
  August/5--8, Ann Arbor, USA: IEEE, 2012, pp. 229--232.

\bibitem{lorbert12}
A.~Lorbert and P.~J. Ramadge, ``Kernel hyperalignment,'' in \emph{25th Advances
  in Neural Information Processing Systems (NIPS)}, December/3--8, Harveys,
  Lake Tahoe, 2012, pp. 1790--1798.

\bibitem{chen14}
P.~H. Chen, J.~S. Guntupalli, J.~V. Haxby, and P.~J. Ramadge, ``Joint
  svd-hyperalignment for multi-subject fmri data alignment,'' in \emph{24th
  International Workshop on Machine Learning for Signal Processing
  (MLSP)}.\hskip 1em plus 0.5em minus 0.4em\relax September/21--24, Reims,
  France: IEEE, 2014, pp. 1--6.

\bibitem{chen15}
P.~H. Chen, J.~Chen, Y.~Yeshurun, U.~Hasson, J.~Haxby, and P.~J. Ramadge, ``A
  reduced-dimension fmri shared response model,'' in \emph{28th Advances in
  Neural Information Processing Systems (NIPS)}, December/7--12, Montréal,
  Canada, 2015, pp. 460--468.

\bibitem{tony16}
M.~Yousefnezhad and D.~Zhang, ``Decoding visual stimuli in human brain by using
  anatomical pattern analysis on fmri images,'' in \emph{8th International
  Conference on Brain Inspired Cognitive Systems (BICS)}.\hskip 1em plus 0.5em
  minus 0.4em\relax November/28--30, Beijing, China: Springer, 2016, pp.
  47--57.

\bibitem{tony17}
M.~Yousefnezhad and D.Zhang, ``Multi-region neural representation: A novel
  model for decoding visual stimuli in human brains,'' in \emph{17th SIAM
  International Conference on Data Mininig (SDM)}, April/27--29, Houston,
  Texas, USA, 2017, pp. 54--62.

\bibitem{tony17a}
M.Yousefnezhad and D.~Zhang, ``Local discriminant hyperalignment for
  multi-subject fmri data alignment,'' in \emph{34th AAAI Conference on
  Artificial Intelligence}, February/4--9, San Francisco, USA, February 2017,
  pp. 59--65.

\bibitem{oswal16}
U.~Oswal, C.~Cox, M.~Lambon-Ralph, T.~Rogers, and R.~Nowak, ``Representational
  similarity learning with application to brain networks,'' in
  \emph{Proceedings of The 33rd International Conference on Machine Learning},
  2016, pp. 1041--1049.

\bibitem{wang2018a}
Y.~Wang, L.~Zhou, B.~Yu, L.~Wang, C.~Zu, D.~S. Lalush, W.~Lin, X.~Wu, J.~Zhou,
  and D.~Shen, ``3d auto-context-based locality adaptive multi-modality gans
  for pet synthesis,'' \emph{IEEE Transactions on Medical Imaging}, vol.~38,
  no.~6, pp. 1328--1339, 2018.

\bibitem{wang2018b}
Y.~Wang, B.~Yu, L.~Wang, C.~Zu, D.~S. Lalush, W.~Lin, X.~Wu, J.~Zhou, D.~Shen,
  and L.~Zhou, ``3d conditional generative adversarial networks for
  high-quality pet image estimation at low dose,'' \emph{NeuroImage}, vol. 174,
  pp. 550--562, 2018.

\bibitem{mohr15}
H.~Mohr, U.~Wolfensteller, S.~Frimmel, and H.~Ruge, ``Sparse regularization
  techniques provide novel insights into outcome integration processes,''
  \emph{NeuroImage}, vol. 104, pp. 163--176, 2015.

\bibitem{talairach88}
J.~Talairach and P.~Tournoux, \emph{Co-planar stereotaxic atlas of the human
  brain. 3-Dimensional proportional system: an approach to cerebral
  imaging}.\hskip 1em plus 0.5em minus 0.4em\relax Thieme, 1988.

\bibitem{mazziotta01}
J.~Mazziotta, A.~Toga, A.~Evans, P.~Fox, J.~Lancaster, K.~Zilles, R.~Woods,
  T.~Paus, G.~Simpson, B.~Pike \emph{et~al.}, ``A probabilistic atlas and
  reference system for the human brain: International consortium for brain
  mapping (icbm),'' \emph{Philosophical Transactions of the Royal Society of
  London B: Biological Sciences}, vol. 356(1412), pp. 1293--1322, 2001.

\bibitem{watson93}
J.~D. Watson, R.~Myers, R.~S.~J. Frackowiak, J.~V. Hajnal, R.~P. Woods, J.~C.
  Mazziotta, S.~Shipp, and S.~Zeki, ``Area v5 of the human brain: evidence from
  a combined study using positron emission tomography and magnetic resonance
  imaging,'' \emph{Cerebral Cortex}, vol.~3, no.~2, pp. 79--94, 1993.

\bibitem{rademacher93}
J.~Rademacher, V.~S. Caviness, H.~Steinmetz, and A.~Galaburda, ``Topographical
  variation of the human primary cortices: implications for neuroimaging, brain
  mapping, and neurobiology,'' \emph{Cerebral Cortex}, vol.~3, no.~4, pp.
  313--329, 1993.

\bibitem{chen16}
P.~Chen, X.~Zhu, H.~Zhang, J.~S. Turek, J.~Chen, T.~Willke, U.~Hasson, and
  P.~Ramadge, ``A convolutional autoencoder for multi-subject fmri data
  aggregation,'' in \emph{29th Workshop of Representation Learning in
  Artificial and Biological Neural Networks.}, NIPS, Dec/5--10, Barcelona,
  2016, pp. 1--9.

\bibitem{conroy09}
B.~Conroy, B.~Singer, J.~Haxby, and P.~J. Ramadge, ``fmri-based inter-subject
  cortical alignment using functional connectivity,'' in \emph{22th Advances in
  Neural Information Processing Systems (NIPS)}, 2009, pp. 378--386.

\bibitem{sabuncu10}
M.~R. Sabuncu, B.~D. Singer, B.~Conroy, R.~E. Bryan, P.~J. Ramadge, and J.~V.
  Haxby, ``Function-based intersubject alignment of human cortical anatomy,''
  \emph{Cerebral Cortex}, vol.~20, no.~1, pp. 130--140, 2010.

\bibitem{michael15}
A.~M. Michael, M.~Anderson, R.~L. Miller, T.~Adal{\i}, and V.~D. Calhoun,
  ``Preserving subject variability in group fmri analysis: performance
  evaluation of gica vs. iva,'' \emph{Distributed Networks-New Outlooks on
  Cerebellar Function}, p. 106, 2015.

\bibitem{Langs10}
G.~Langs, Y.~Tie, L.~Rigolo, A.~Golby, and P.~Golland, ``Functional geometry
  alignment and localization of brain areas,'' in \emph{Advances in Neural
  Information Processing Systems (NIPS)}, 2010, pp. 1225--1233.

\bibitem{schonemann66}
P.~H. Sch{\"o}nemann, ``A generalized solution of the orthogonal procrustes
  problem,'' \emph{Psychometrika}, vol.~31, no.~1, pp. 1--10, 1966.

\bibitem{gower04}
J.~C. Gower and G.~B. Dijksterhuis, \emph{Procrustes problems}.\hskip 1em plus
  0.5em minus 0.4em\relax Oxford University Press on Demand, 2004, vol.~30.

\bibitem{dmochowski12}
J.~P. Dmochowski, P.~Sajda, J.~Dias, and L.~C. Parra, ``Correlated components
  of ongoing eeg point to emotionally laden attention--a possible marker of
  engagement?'' \emph{Frontiers in Human Neuroscience}, vol.~6, p. 112, 2012.

\bibitem{Tony17NIPS}
M.~Yousefnezhad and D.~Zhang, ``Deep hyperalignment,'' in \emph{Advances in
  Neural Information Processing Systems (NIPS)}, 2017, pp. 1603--1611.

\bibitem{sui11}
J.~Sui, G.~Pearlson, A.~Caprihan, T.~Adali, K.~A. Kiehl, J.~Liu, J.~Yamamoto,
  and V.~D. Calhoun, ``Discriminating schizophrenia and bipolar disorder by
  fusing fmri and dti in a multimodal cca+ joint ica model,''
  \emph{NeuroImage}, vol.~57, no.~3, pp. 839--855, 2011.

\bibitem{sui13}
J.~Sui, H.~He, G.~D. Pearlson, T.~Adali, K.~A. Kiehl, Q.~Yu, V.~P. Clark,
  E.~Castro, T.~White, B.~A. Mueller \emph{et~al.}, ``Three-way (n-way) fusion
  of brain imaging data based on mcca+ jica and its application to
  discriminating schizophrenia,'' \emph{NeuroImage}, vol.~66, pp. 119--132,
  2013.

\bibitem{guntupalli16}
J.~S. Guntupalli, M.~Hanke, Y.~O. Halchenko, A.~C. Connolly, P.~J. Ramadge, and
  J.~V. Haxby, ``A model of representational spaces in human cortex,''
  \emph{Cerebral Cortex}, p. bhw068, 2016.

\bibitem{turek16}
J.~S. Turek, T.~L. Willke, P.-H. Chen, and P.~J. Ramadge, ``A semi-supervised
  method for multi-subject fmri functional alignment,'' in \emph{41st
  International Conference on Acoustics, Speech and Signal Processing
  (ICASSP)}.\hskip 1em plus 0.5em minus 0.4em\relax March/20--25, Shanghai,
  China: IEEE, 2016, pp. 1--5.

\bibitem{rastogi15}
P.~Rastogi, B.~Van~Durme, and R.~Arora, ``Multiview lsa: Representation
  learning via generalized cca.'' in \emph{14th Annual Conference of the North
  American Chapter of the Association for Computational Linguistics: Human
  Language Technologies (HLT-NAACL)}, May/31 to June/5, Denver, Colorado, USA,
  2015, pp. 556--566.

\bibitem{benton17}
A.~Benton, H.~Khayrallah, B.~Gujral, D.~Reisinger, S.~Zhang, and R.~Arora,
  ``Deep generalized canonical correlation analysis,'' in \emph{5th
  International Conference on Learning Representations (ICLR)}, April/24--26,
  Toulon, France, 2017, pp. 1--14.

\bibitem{arora12}
R.~Arora and K.~Livescu, ``Kernel cca for multi-view learning of acoustic
  features using articulatory measurements.'' in \emph{Machine Learning in
  Speech and Language Processing (MLSLP)}.\hskip 1em plus 0.5em minus
  0.4em\relax September/14, Portland, Oregon, USA: Citeseer, 2012, pp. 34--37.

\bibitem{brand02}
M.~Brand, ``Incremental singular value decomposition of uncertain data with
  missing values,'' in \emph{7th European Conference on Computer Vision
  (ECCV)}.\hskip 1em plus 0.5em minus 0.4em\relax May/28--31, Copenhagen,
  Denmark: Springer, 2002, pp. 707--720.

\bibitem{tom07}
S.~M. Tom, C.~R. Fox, C.~Trepel, and R.~A. Poldrack, ``The neural basis of loss
  aversion in decision-making under risk,'' \emph{Science}, vol. 315, no. 5811,
  pp. 515--518, 2007.

\bibitem{duncan09}
K.~J. Duncan, C.~Pattamadilok, I.~Knierim, and J.~T. Devlin, ``Consistency and
  variability in functional localisers,'' \emph{NeuroImage}, vol.~46, no.~4,
  pp. 1018--1026, 2009.

\bibitem{hanke14}
M.~Hanke, F.~J. Baumgartner, P.~Ibe, F.~R. Kaule, S.~Pollmann, O.~Speck,
  W.~Zinke, and J.~Stadler, ``A high-resolution 7-tesla fmri dataset from
  complex natural stimulation with an audio movie,'' \emph{Scientific Data},
  vol.~1, 2014.

\bibitem{walz13}
J.~M. Walz, R.~I. Goldman, M.~Carapezza, J.~Muraskin, T.~R. Brown, and
  P.~Sajda, ``Simultaneous eeg-fmri reveals temporal evolution of coupling
  between supramodal cortical attention networks and the brainstem,''
  \emph{Journal of Neuroscience}, vol.~33, no.~49, pp. 19\,212--19\,222, 2013.

\bibitem{wakeman15}
D.~G. Wakeman and R.~N. Henson, ``A multi-subject, multi-modal human
  neuroimaging dataset,'' \emph{Scientific Data}, vol.~2, 2015.

\bibitem{mitchell08}
T.~M. Mitchell, S.~V. Shinkareva, A.~Carlson, K.-M. Chang, V.~L. Malave, R.~A.
  Mason, and M.~A. Just, ``Predicting human brain activity associated with the
  meanings of nouns,'' \emph{science}, vol. 320, no. 5880, pp. 1191--1195,
  2008.

\bibitem{song16}
X.~Song and H.~Lu, ``Multilinear regression for embedded feature selection with
  application to fmri analysis,'' in \emph{31st AAAI Conference on Artificial
  Intelligence}.\hskip 1em plus 0.5em minus 0.4em\relax Association for the
  Advancement of Artificial Intelligence, 2016, pp. 2562--2568.

\bibitem{jenkinson01}
M.~Jenkinson and S.~Smith, ``A global optimisation method for robust affine
  registration of brain images,'' \emph{Medical Image Analysis}, vol.~5, no.~2,
  pp. 143--156, 2001.

\bibitem{jenkinson02}
M.~Jenkinson, P.~Bannister, M.~Brady, and S.~Smith, ``Improved optimization for
  the robust and accurate linear registration and motion correction of brain
  images,'' \emph{NeuroImage}, vol.~17, no.~2, pp. 825--841, 2002.

\bibitem{smith02}
S.~M. Smith, ``Fast robust automated brain extraction,'' \emph{Human Brain
  Mapping}, vol.~17, no.~3, pp. 143--155, 2002.

\bibitem{woolrich01}
M.~W. Woolrich, B.~D. Ripley, M.~Brady, and S.~M. Smith, ``Temporal
  autocorrelation in univariate linear modeling of fmri data,''
  \emph{Neuroimage}, vol.~14, no.~6, pp. 1370--1386, 2001.

\bibitem{smola04}
A.~J. Smola and B.~Sch{\"o}lkopf, ``A tutorial on support vector regression,''
  \emph{Statistics and Computing}, vol.~14, no.~3, pp. 199--222, 2004.

\bibitem{Anderson16}
M.~J. Anderson, M.~Capota, J.~S. Turek, X.~Zhu, T.~L. Willke, Y.~Wang, P.-H.
  Chen, J.~R. Manning, P.~J. Ramadge, and K.~A. Norman, ``Enabling factor
  analysis on thousand-subject neuroimaging datasets,'' in \emph{2016 IEEE
  International Conference on Big Data (Big Data)}.\hskip 1em plus 0.5em minus
  0.4em\relax IEEE, 2016, pp. 1151--1160.

\end{thebibliography}

\vskip -0.8in
\begin{IEEEbiography}[{\includegraphics[width=1in,height=1.25in,clip,keepaspectratio]{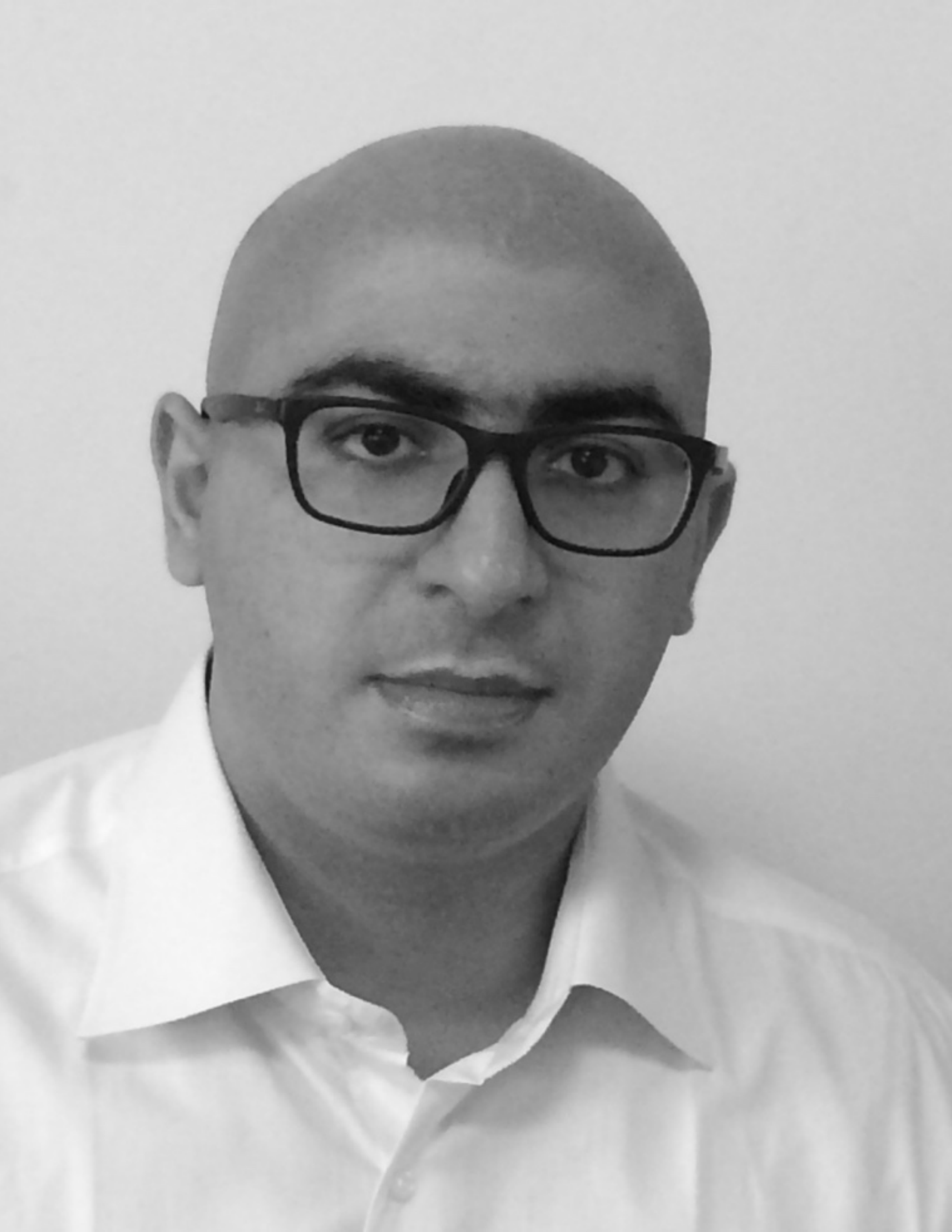}}]{Muhammad~Yousefnezhad} received his B.Sc. and M.Sc. degrees in Computer Hardware Engineering and Information Technology (IT), with a sheer focus on Artificial Intelligence, from Mazandaran University of Science and Technology, Iran, in 2010 and 2013, respectively. He received his Ph.D. in the Department of Computer Science and Technology at Nanjing University of Aeronautics and Astronautics (NUAA), China, in 2018. He is now a Postdoctoral Fellow working with the Department of Computing Science and the Department of Psychiatry at the University of Alberta, Canada, since March 2019. His primary research interest lies in developing machine learning techniques, particularly within the area of the human brain decoding.
\end{IEEEbiography}
\vskip -0.8in
\begin{IEEEbiography}[{\includegraphics[width=1in,height=1.25in,clip,keepaspectratio]{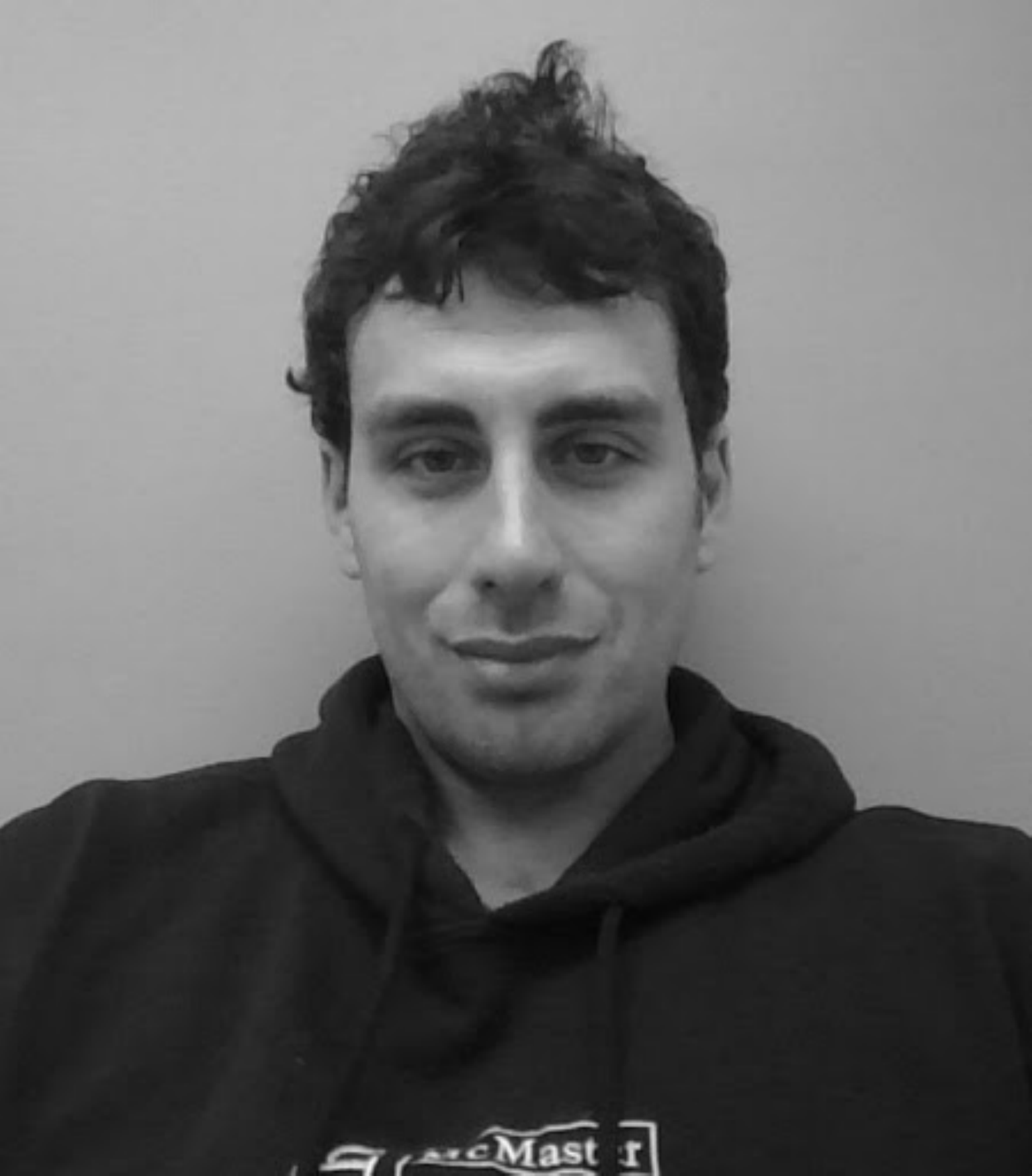}}]{Alessandro~Selvitella} received a BSc and an MSc in Mathematics from Universita' degli Studi di Milano (Milan, Italy), a PhD in Mathematical Analysis from SISSA (Trieste, Italy),  an MSc and a PhD in Statistics from McMaster University (Hamilton, Canada). Alessandro has been a Postdoctoral Fellow in Applied Mathematics at McMaster University and he has been a Postdoctoral Fellow in the Departments of Psychiatry and Computing Science at the University of Alberta. Alessandro is currently an Assistant Professor in the Department of Mathematical Sciences at Purdue University (Fort Wayne, U.S.). Alessandro' current research interests are in the theory and applications of Machine Learning methods to the Biological and Medical Sciences.
\end{IEEEbiography}
\vskip -4.3in
\begin{IEEEbiography}[{\includegraphics[width=1in,height=1.25in,clip,keepaspectratio]{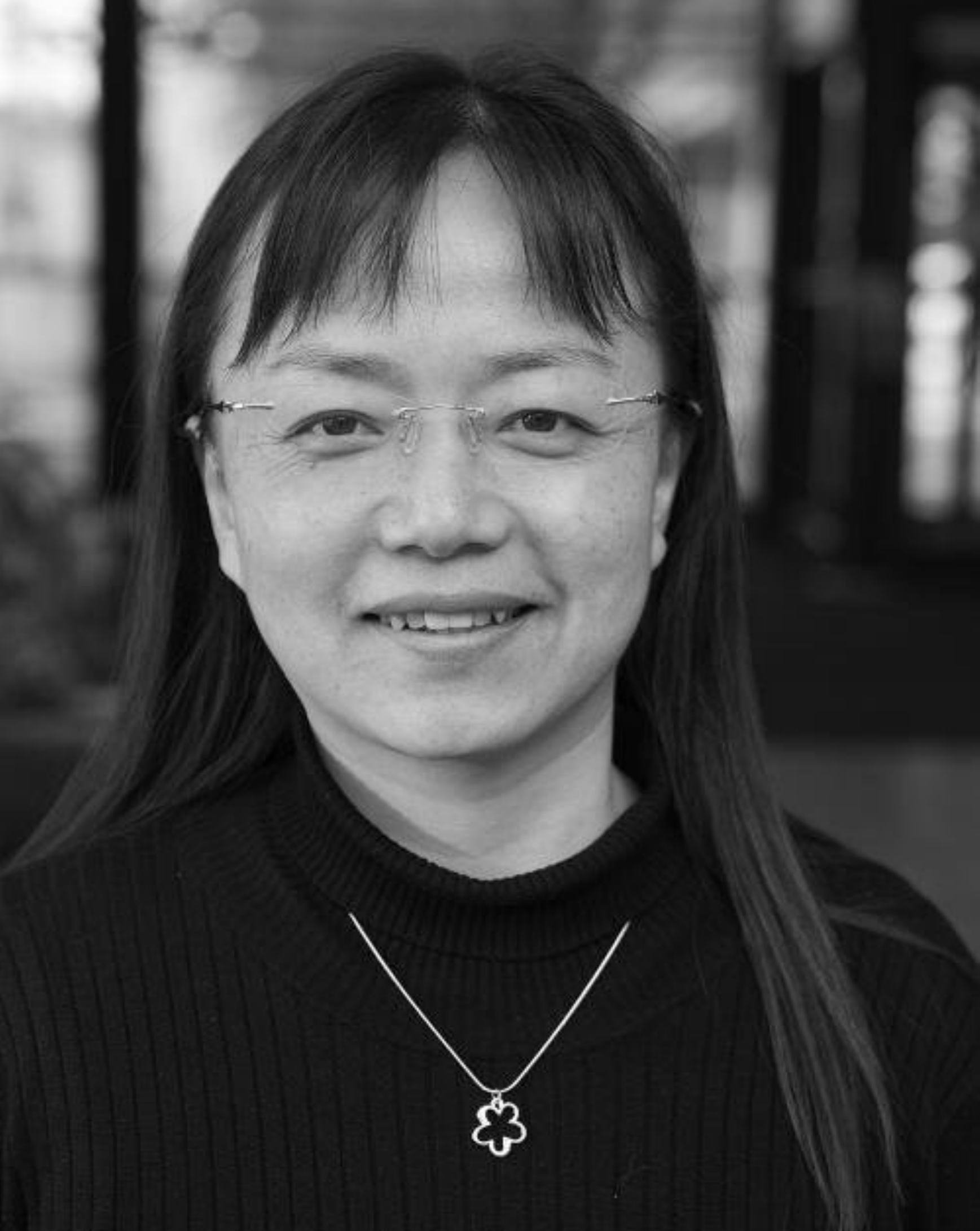}}]{Liangxiu~Han} received the Ph.D. degree in computer science from Fudan University, Shanghai, China, in 2002. She is currently a Full Professor of computer science with the School of Computing, Mathematics, and Digital Technology, Manchester Metropolitan University.  She is also the Deputy Director of the Center for Advanced Computational Science and Man Met Crime and Well-Being Big Data Center. Her research areas mainly lie in the development of novel big data analytics and development of novel intelligent architectures that facilitates big data analytics (e.g., parallel and distributed computing, Cloud/Service-oriented computing/data intensive computing) as well as applications in different domains using various large datasets (biomedical images, environmental sensor, network traffic data, web documents, etc.). She is currently a Principal Investigator or Co-PI on a number of research projects in the research areas mentioned above.
\end{IEEEbiography}
\vskip -4.3in
\begin{IEEEbiography}[{\includegraphics[width=1in,height=1.25in,clip,keepaspectratio]{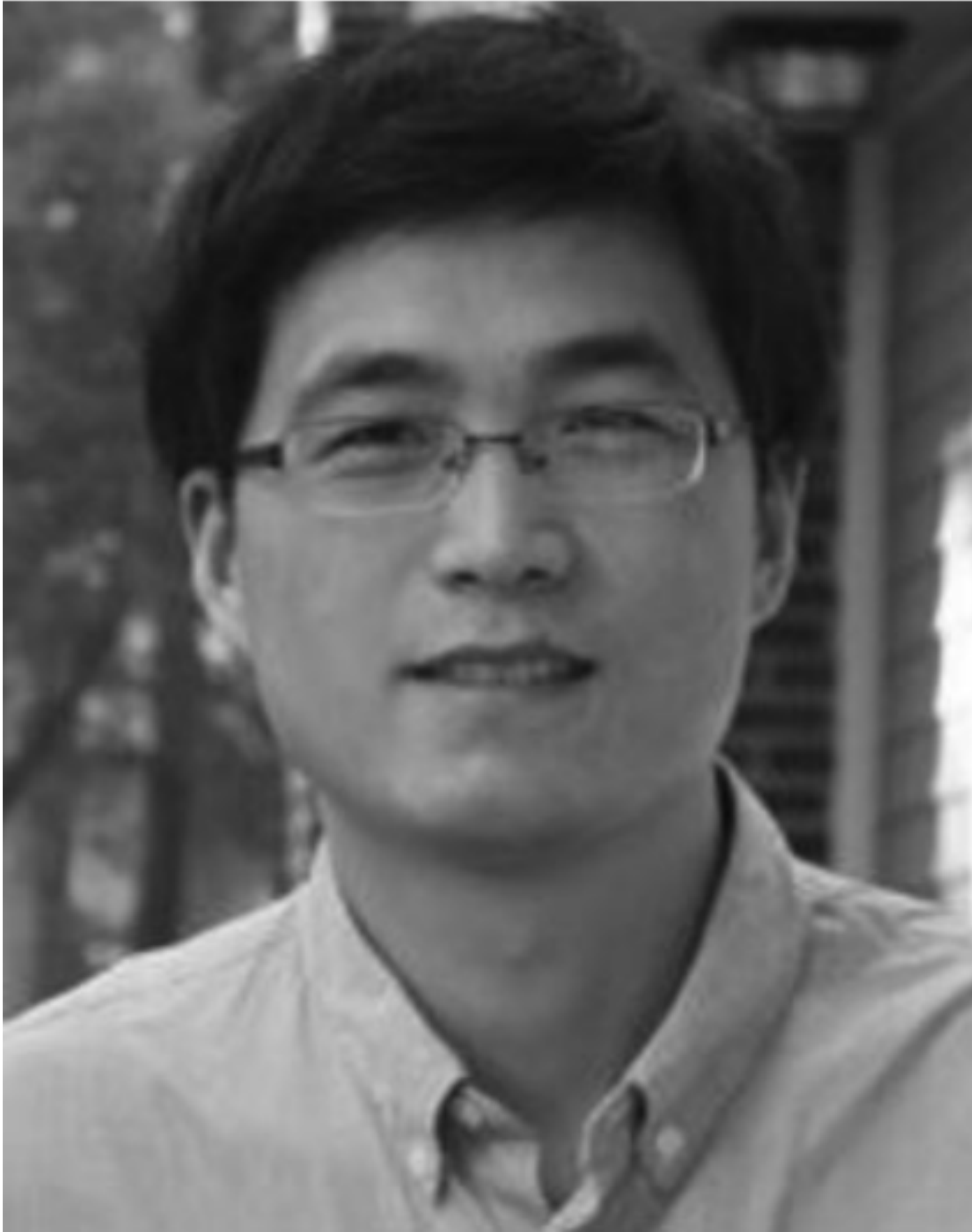}}]{Daoqiang~Zhang}
	received the B.Sc. and Ph.D. degrees in computer science from Nanjing University of Aeronautics and Astronautics, Nanjing, China, in 1999 and 2004, respectively. He is currently a Professor in the Department of Computer Science and Engineering, Nanjing University of Aeronautics and Astronautics. His current research interests include machine learning, pattern recognition, and biomedical image analysis. In these areas, he has authored or coauthored more than 100 technical papers in the refereed international journals and conference proceedings. 
\end{IEEEbiography}

\end{document}